\documentclass[10pt,journal,compsoc]{IEEEtran}
\ifCLASSOPTIONcompsoc
  \usepackage[nocompress]{cite}
\else
  \usepackage{cite}
  
\fi

\ifCLASSINFOpdf
\else
\fi

\usepackage{epsfig}
\usepackage{graphicx}
\usepackage{amsmath}
\usepackage{amssymb}

\usepackage{cases}
\usepackage{subfigure}
\usepackage{multirow}
\usepackage{mathrsfs}
\usepackage{bm}
\usepackage{amsthm}
\usepackage{mathtools}
\usepackage{color}
\usepackage{url}
\usepackage[ruled]{algorithm2e}

\newcounter{thm_counter}

\newtheorem{theorem}[thm_counter]{Theorem}

\begin{document}
%
\title{ATZSL: Defensive Zero-Shot Recognition in the Presence of Adversaries}


\author{Xingxing Zhang,
        Shupeng Gui,
        Zhenfeng Zhu,
        Yao Zhao,~\IEEEmembership{Senior Member,~IEEE,}
        and~Ji~Liu
\IEEEcompsocitemizethanks{\IEEEcompsocthanksitem X. Zhang, Z. Zhu, and Y. Zhao are with the Institute of Information Science, Beijing Jiaotong University, Beijing 100044, China, and also with the Beijing Key Laboratory of Advanced Information Science and Network Technology, Beijing Jiaotong University, Beijing 100044, China (e-mail: zhangxing@bjtu.edu.cn; zhfzhu@bjtu.edu.cn; yzhao@bjtu.edu.cn). (\emph {Corresponding author: Yao Zhao.}) \protect\\
\IEEEcompsocthanksitem S. Gui and J. Liu are with the Department of Computer Science, University of Rochester, Rochester, NY 14611 USA (e-mail: shupenggui@gmail.com; ji.liu.uwisc@gmail.com).}}


\IEEEtitleabstractindextext{%
\begin{abstract}
Zero-shot learning (ZSL) has received extensive attention recently especially in areas of fine-grained object recognition, retrieval, and image captioning. Due to the complete lack of training samples and high requirement of defense transferability, the ZSL model learned is particularly vulnerable against adversarial attacks. Recent work also showed adversarially robust generalization requires more data. This may significantly affect the robustness of ZSL. However, very few efforts have been devoted towards this direction. In this paper, we take an initial attempt, and propose a generic formulation to provide a systematical solution (named \textbf{ATZSL}) for learning a robust ZSL model. It is capable of achieving better generalization on various adversarial objects recognition while only losing a negligible performance on clean images for unseen classes, by casting ZSL into a min-max optimization problem. To address it, we design a defensive relation prediction network, which can bridge the seen and unseen class domains via attributes to generalize prediction and defense strategy. Additionally, our framework can be extended to deal with the poisoned scenario of unseen class attributes. An extensive group of experiments are then presented, demonstrating that ATZSL obtains remarkably more favorable trade-off between model transferability and robustness, over currently available alternatives under various settings.
\end{abstract}

\begin{IEEEkeywords}
Zero-shot learning, robustness, optimization, defensive, unseen class
\end{IEEEkeywords}}

\maketitle

\IEEEdisplaynontitleabstractindextext

%
\IEEEpeerreviewmaketitle

\IEEEraisesectionheading{\section{Introduction}\label{sec:introduction}}

\IEEEPARstart{I}{n} many practical applications, we need the model to have the ability to determine the class labels for the data belonging to unseen classes. The following are some popular application scenarios~\cite{wang2019survey}:
 \begin{itemize}
     \item[-] \emph{The number of target classes is large}. Generally, human beings can recognize at least 30,000 object classes~\cite{biederman1987recognition}. However, collecting sufficient labeled instances for such a large number of classes is challenging. Thus, existing image datasets can only cover a small subset of these classes.
     \item[-] \emph{Target classes are rare}. An example is fine-grained object recognition. Suppose we want to recognize flowers of different breeds~\cite{lei2015predicting}. It is hard and even prohibitive to collect sufficient image instances for each specific flower breed. 
     \item[-] \emph{Target classes change over time}. 
     An example is recognizing images of products belonging to a certain style and brand. As products of new styles and new brands appear frequently, for some new products, it is difficult to find corresponding labeled instances~\cite{long2017zero}.
     \item[-] \emph{Annotating instances is expensive and time consuming}. For example, in the image captioning problem~\cite{venugopalan2017captioning}, each image in the training data should have a corresponding caption. This problem can be seen as a sequential classification problem. The number of object classes covered by the existing image-text corpora is limited, with many classes not being covered in practice.
 \end{itemize}
 
\begin{figure}[t!]
\begin{center}
\centerline{\includegraphics[width=3.3in]{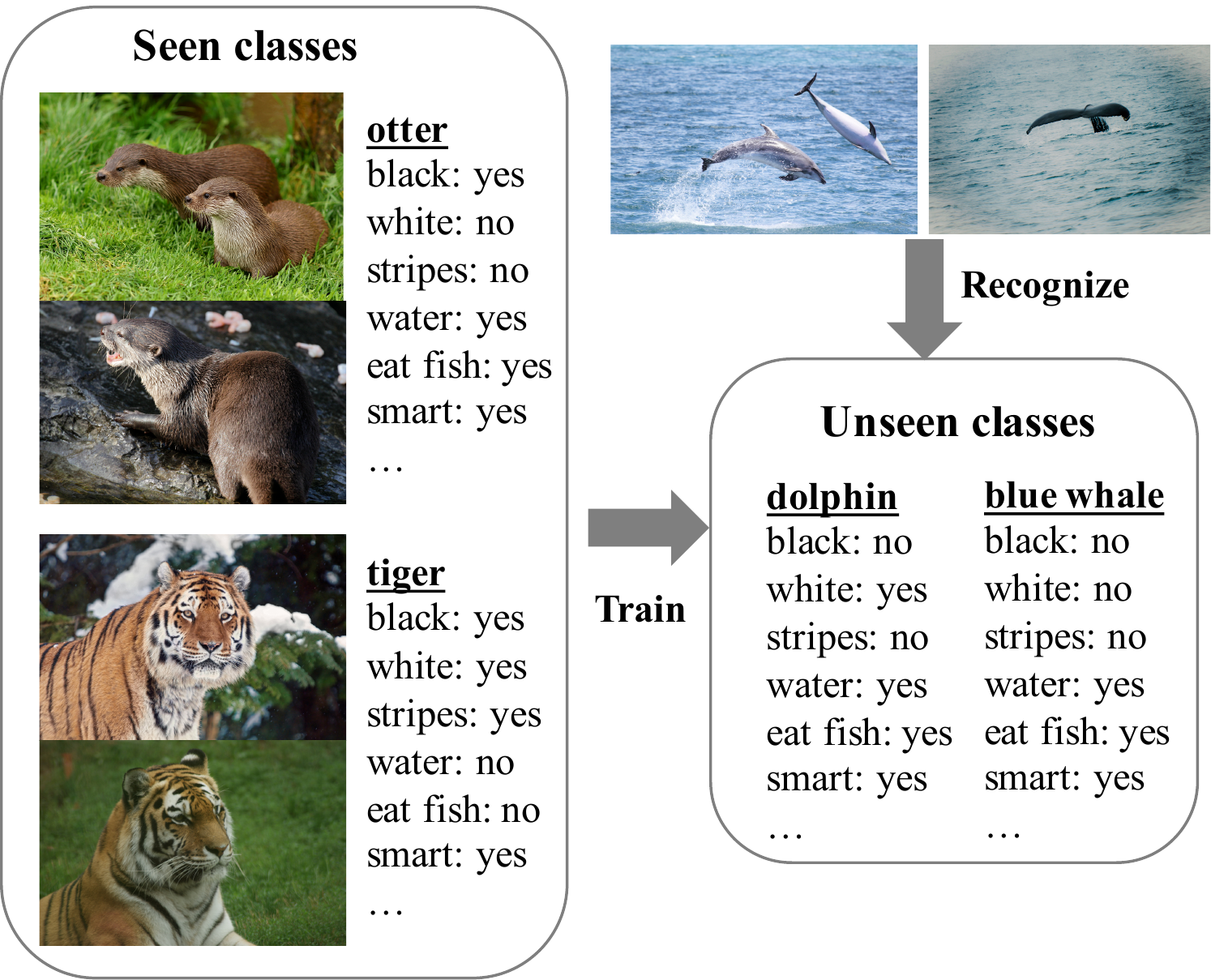}}
\caption{The visual images and class prototypes provided for several classes in benchmark dataset AWA2~\cite{xian2018zero}.}
\label{att_image}
\end{center}
\end{figure}
 
 To solve this problem, Zero-Shot Learning (ZSL)~\cite{mensink2014costa,zhang2015zero, changpinyo2016synthesized,morgado2017semantically,xian2018feature,felix2018multi,schonfeld2019generalized,kampffmeyer2019rethinking} is proposed. The goal is to recognize data belonging to the classes that have no labeled samples. Since its inception, it has become a fast-developing field in machine learning with a wide range of applications in computer vision. Due to lack of labeled samples in the unseen class domain, auxiliary information is necessary for ZSL to transfer knowledge from the seen to the unseen classes.  
 As shown in Fig.~\ref{att_image}, existing methods usually provide each class with one \emph{class prototype derived from text} (e.g., attribute vector~\cite{kankuekul2012online, lampert2009learning, lampert2014attribute} or word vector~\cite{akata2015evaluation, frome2013devise, socher2013zero}). 
 This is inspired by the way human beings recognize the world. For example, with the knowledge that ``a zebra looks like a horse, and with stripes'', we can recognize a zebra even without having seen one before, as long as we know what a ``horse'' is and how ``stripes'' look like.

 Despite achieving outstanding performance on general recognition tasks, state-of-the-art classifiers remain highly vulnerable to small, imperceptible, and adversarial perturbations~\cite{fawzi2018adversarial}. This vulnerability in the ZSL scenario is also inevitable, where an unseen object is transformed by an undetectable amount. 
 It may be even easier to be attacked for ZSL models compared with normal supervised learning models, since~\cite{schmidt2018adversarially} has proved that adversarially robust generalization requires more data, while there does not exist any training data for unseen classes in ZSL.
 In particular, while people mostly care about the transferability and discriminability, the robustness of ZSL models is usually ignored though it is also extremely important.
 We have empirically proved the adversarial perturbations in zero-shot recognition are more intricate to be addressed by existing ZSL models. This comes from the fact that without labeled samples in the unseen class domain, ZSL models need to transfer knowledge not only from the seen to the unseen classes, but also from the clean to the adversarial examples with unknown adversaries. Consequently, it is difficult and risky to apply such fragile ZSL models in practice, especially in security-critical areas. For example, if we use a zero-shot activity recognition model in self-driving cars, adversarial examples of those unseen activities could allow an attacker to cause the car to take unwanted actions. 
 
 Besides in the visual space, the adversarial examples in the semantic space are also inevitable. This is because the class prototypes of unseen classes, either human-defined attributes or automatically extracted word vectors, are all obtained independently from visual samples, and then easily been poisoned undetectably. However, these prototypes play a key role in transferring knowledge from the seen to the unseen classes, and thus even a small perturbation will degrade the recognition performance. One example, when the description is changed to ``a zebra looks like a horse, and \emph{without} stripes'', we are not able to recognize a zebra ``without stripes''.
 
 Motivated by the two key observations in the visual and semantic spaces above, in this work, we focus on developing a robust ZSL model. 
 Different from previous ZSL methods, we propose to train a novel ZSL model adversarially. By injecting adversarial examples into our model, it is able to achieve promising generalization on unseen objects recognition under various image or attribute attacks. And meanwhile, the recognition performance on clean data has only non-obvious degradation by designing a defensive relation prediction network.
 In particular, the added overhead to generate and train adversarial examples can be negligible.
 We emphasize our \textbf{contributions} in four aspects:
\begin{itemize}
\item[-] To the best of our knowledge, our work describes the \underline{first algorithmic framework} to jointly optimize the two key goals: ZSL robustness against adversarial attacks, and transferability to unseen classes.
\item[-] We develop ATZSL: an \underline{A}dversarially \underline{T}rained \underline{Z}ero-\underline{S}hot \underline{L}earning model, to alleviate the instability issue of ZSL caused by small, well sought perturbations of both images and attributes. 
\item[-] By casting ZSL into a constrained min-max optimization form, our ATZSL transfers knowledge from the clean to the adversarial examples, thus achieving competitive trade-off between the performances in clean and adversarial domains.
\item[-] The seen and unseen class domains are bridged by a defensive relation prediction network in ATZSL. The improvements over alternative ZSL models are consistently significant under various settings.
\end{itemize}

\begin{table*}[!t]
\renewcommand{\arraystretch}{1.3}
\caption{Key Notations} 
 \vspace{-0.3cm}
 \label{tab:notation}
\centering
   \begin{tabular}{c|l}
    \hline 
     \textbf{Notation}  & \textbf{Description}       \\
     \hline
     $\mathcal{S},\mathcal{U}$&Set of seen classes and set of unseen classes, respectively\\
     \hline
     $\mathcal{X},\mathcal{P}$&Visual space and semantic space, respectively\\ \hline
    $N_{\textrm{tr}}$ & Number of training samples\\ \hline
     $N_{\textrm{s}},N_{\textrm{u}}$ & Number of seen classes and number of unseen classes, respectively\\ \hline
     $(x_{i}^{\textrm{tr}},y_{i}^{\textrm{tr}})$ & The $i$th labeled training sample: image $x_{i}^{\textrm{tr}}\in \mathcal{X} $ and label $y_{i}^{\textrm{tr}}\in\mathcal{S}$\\ \hline
     $x_{i}^{\textrm{te}}$ & The $i$th unlabeled testing sample: image $x_{i}^{\textrm{te}}\in \mathcal{X}$\\ \hline
     $(c_{i}^{\textrm{s}},\bm p_{i}^{\textrm{s}})$ & The $i$th seen class $c_{i}^{\textrm{s}}\in\mathcal{S}$ and its class prototype $\bm p_{i}^{\textrm{s}}\in\mathcal{P}$\\ \hline
     $(c_{i}^{\textrm{u}},\bm p_{i}^{\textrm{u}})$ & The $i$th unseen class $c_{i}^{\textrm{u}}\in\mathcal{U}$ and its class prototype $\bm p_{i}^{\textrm{u}}\in\mathcal{P}$\\\hline
     $x_{i}^{{\textrm{tr}}'},x_{i}^{{\textrm{te}}'}$ & The $i$th training sample and testing sample both with an adversarial perturbation\\ \hline     
     $\bm p_{i}^{{\textrm{s}}'},\bm p_{i}^{{\textrm{u}}'}$ & The $i$th seen class prototype and unseen class prototype both with an adversarial perturbation\\\hline
     $\rho$ & Attack magnitude\\
    \hline
    \end{tabular}
\end{table*}

\section{Related Work}\label{Related work}

In this section, we first briefly introduce some related works of zero-shot learning, and then present a review for various adversarial attacks.

\textbf{Zero-shot Learning.} According to whether information about the testing data is involved during model learning, existing ZSL models consist of inductive~\cite{liu2019convolutional,annadani2018preserving, changpinyo2016synthesized, kodirov2017semantic, romera2015embarrassingly} and transductive settings~\cite{fu2015transductive,fu2018zero,kodirov2015unsupervised,niu2018webly,zhang2019hierarchical}. Specifically, this transduction in ZSL can be embodied in two progressive degrees: transductive for specific unseen classes~\cite{liu2018generalized} and transductive for specific testing samples~\cite{zhao2018domain}. Although transductive settings can rectify the domain shift caused by the different distributions of the training and the testing samples, it is problematic in many practical scenarios. Thus, we take an inductive setting in this work.

From the view of constructing the visual-semantic interactions, existing inductive ZSL methods fall into four categories. 
The first group learns a projection function from the visual to the semantic space with a linear~\cite{bansal2018zero, lampert2014attribute, li2018discriminative} or a non-linear model~\cite{chen2018zero,  morgado2017semantically, socher2013zero, yu2018stacked}. The testing unseen data are then classified by matching the visual representations in the class semantic embedding space with the unseen class prototypes. 
The second group chooses the reverse projection direction~\cite{annadani2018preserving,wang2018zero,zhang2017learning}, to alleviate the hubness problem caused by nearest neighbour search in a high dimensional space~\cite{radovanovic2010hubs}. The testing unseen data are then classified by resorting to the most similar pseudo visual examples in the visual space. 
The third group is a combination of the first two groups by taking the encoder-decoder paradigm but with the visual feature or class prototype reconstruction constraint~\cite{annadani2018preserving, kodirov2017semantic}. It has been verified that the projection function learned in this way is able to generalize better to the unseen classes. 
The last group learns an intermediate space, where both the visual and the semantic spaces are projected to~\cite{changpinyo2018classifier, changpinyo2017predicting, hubert2017learning,liu2018generalized}. 

\begin{figure}[t!]
\begin{center}
\centerline{\includegraphics[width=\columnwidth]{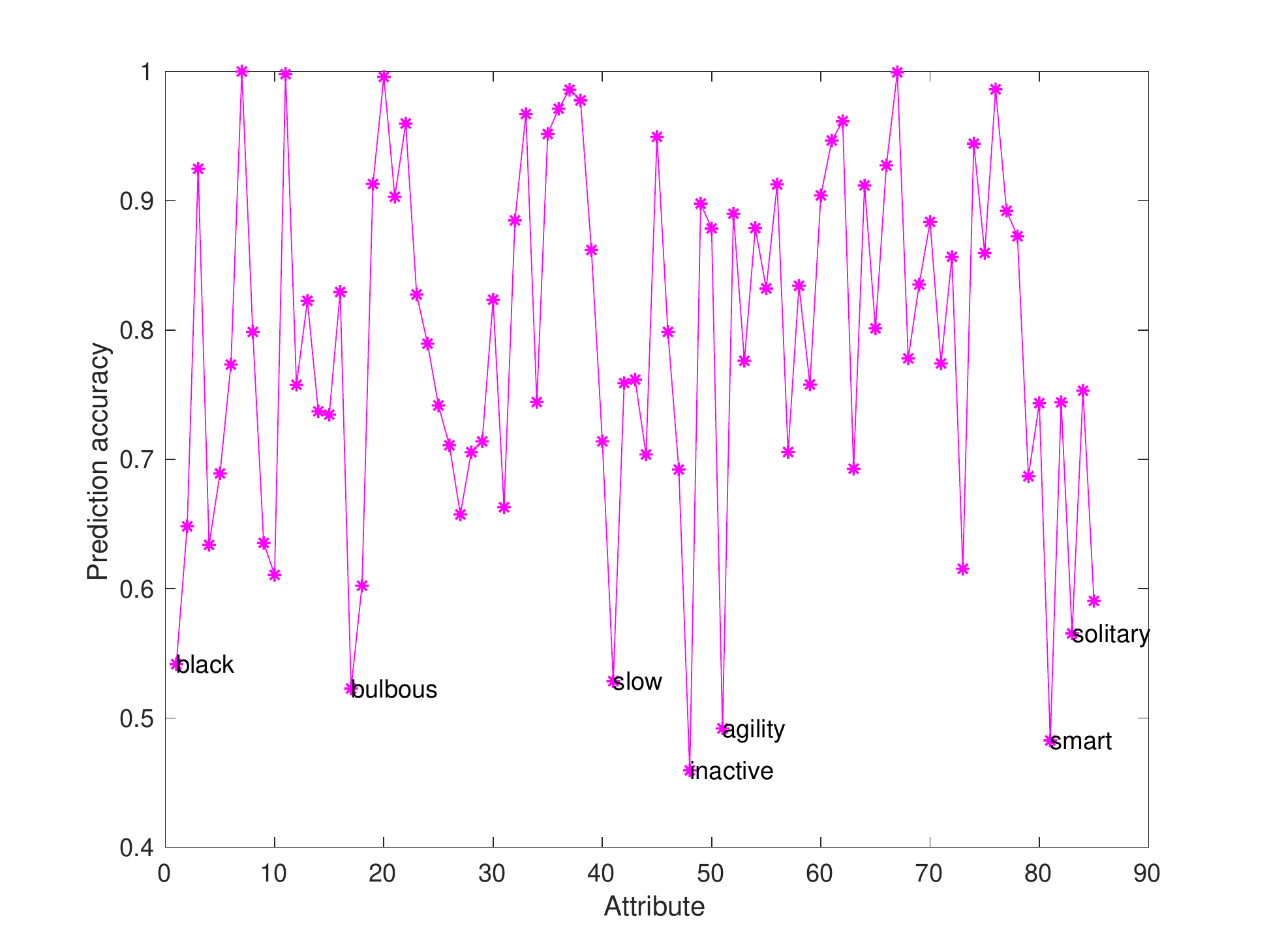}}
\caption{The predictability of each binary attribute, measured with classification accuracy by fine-tuning the last layer of Resnet101~\cite{he2016deep}.}
\label{error}
\end{center}
\end{figure}

 However, they all ignore a fact that each class in ZSL is only provided with one attribute vector, which is not enough to represent all the samples in this class. 
 Consequently, the learned projection may not be effective enough to recognize the samples from the same class.
 Besides, the non-visual components are often included in the provided semantic prototypes, such as ``smart'', ``agility'', and ``inactive'' in benchmark dataset AWA2~\cite{xian2018zero}. Based only on visual information, these attributes are almost impossible to be predicted and even with the level of random guess as observed from Fig.~\ref{error}. Thus, the learned projection cannot generalize well on the unseen class domain, although work normally in the seen class domain through supervised training.
 Considering such facts, we choose to learn an intermediate space to perform zero-shot recognition. In particular, different from existing models, our ATZSL connects the visual and semantic embeddings in series instead of in parallel to avoid amplifying the small perturbations.

\textbf{Adversarial Attacks.} 
Evaluating the robustness of a neural network can be done by crafting adversarial examples at the testing time with a specific attack algorithm~\cite{carlini2017towards, chen2017zoo, goodfellow2014explaining,kurakin2016adversarial,madry2017towards, moosavi2016deepfool, sinha2017certifying}. 
It is worth noting that adversarial attacks do not tamper with the targeted model but rather force it to produce incorrect outputs. The effectiveness of such attacks is determined mainly by the amount of information available to the adversary about the model. There is a large body of work on testing phase attacks, it can be broadly classified into either \emph{White-Box} or \emph{Black-Box} attacks. An adversary in white-box attack has total knowledge about the model used for inference. The access to internal model weights for a white-box attack, such as FGSM~\cite{goodfellow2014explaining} and Deepfool~\cite{moosavi2016deepfool}, usually corresponds to a very strong adversarial attack since it works on the gradient of the network loss function. Black-box attack (e.g., ZOO~\cite{chen2017zoo} and NES-PGD~\cite{ilyas2018black}), on the contrary, assumes no knowledge about the targeted model. It usually analyses the vulnerability of the model by generating an implicit approximation to the actual gradient, based on a greedy local search. 

\section{Adversarially Trained Zero-Shot Learning}\label{Method}
In this section, we first set up the ZSL problem in the presence of adversaries~(Section~\ref{Setup}), then develop an ATZSL model to improve robustness of ZSL in the visual space~(Section~\ref{ModelFormulation}), and finally derive an algorithm to solve ATZSL~(Section~\ref{ModelOptimization}).


\subsection{Problem Definition}\label{Setup}
Given labeled training samples ${\mathcal{D}}^{\textrm{tr}}=\left\{(x_{i}^{\textrm{tr}},y_{i}^{\textrm{tr}})\right\}_{i=1}^{N_{\textrm{tr}}}$ belonging to the set of seen classes $\mathcal{S}$, ATZSL aims to learn a robust zero-shot classifier $f^{u}(\cdot ):\mathcal{X}\rightarrow\mathcal{U}$ that is effective especially in the following two scenarios: i) ATZSL can classify not only a clean testing sample $x_{i}^{\textrm{te}}$ but also its adversarial sample $x_{i}^{{\textrm{te}}'}$ belonging to the set of unseen classes $\mathcal{U}$; ii) ATZSL can classify a clean testing sample $x_{i}^{\textrm{te}}$ even when the unseen class prototypes are attacked. The key notations used throughout this paper are summarized in Table~\ref{tab:notation}.

\subsection{ATZSL: Formulation}\label{ModelFormulation}
\begin{figure*}[htpb!]
    \centering
    \includegraphics[width=0.9\linewidth]{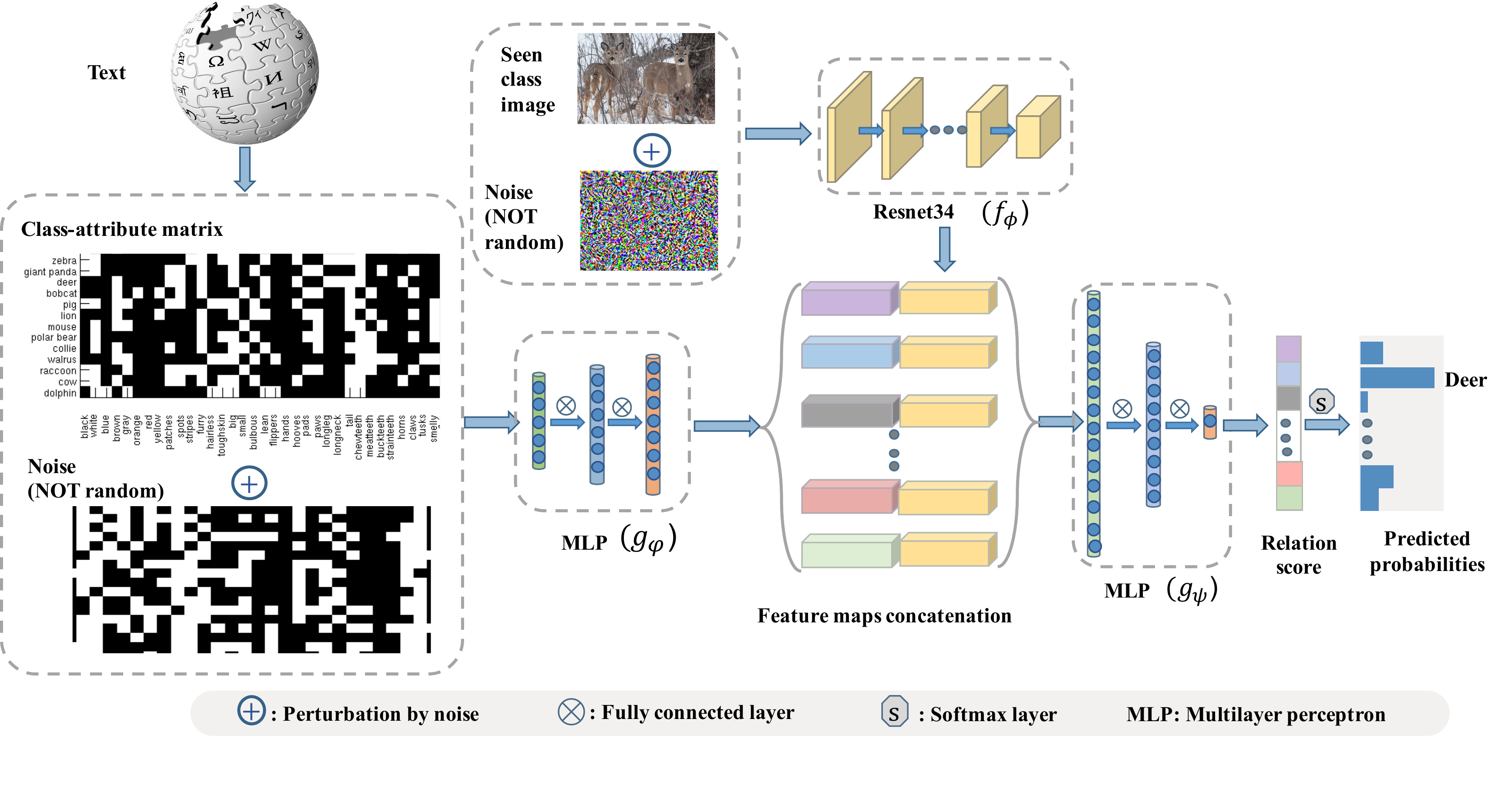}
    \caption{The architecture of the proposed ATZSL to perform robust zero-shot recognition with a defensive relation prediction network. $f_{\phi}$, $g_{\varphi}$, and $g_{\psi}$ represent the feature extraction module, attribute embedding module, and relation prediction module, respectively. }
    \label{fig:Framework}
\end{figure*}

The key to ZSL is to transfer knowledge from the seen to the unseen classes. Motivated by fact that class prototypes (e.g., attribute vectors) can bridge the two domains, we consider transferring the prediction strategy of image-attribute relation.
For this end, we design a relation prediction network that consists of three modules: a feature extraction module $f_{\phi}$, an attribute embedding module $g_{\varphi}$, and a relation prediction module $g_{\psi}$, as illustrated in Fig.~\ref{fig:Framework}. The image $x_{i}^{\textrm{tr}}$ and the attribute $\bm p_{j}^{\textrm{s}}$ are fed into the feature extraction module $f_{\phi}$ and the attribute embedding module $g_{\varphi}$, respectively, which produce two embedding vectors $f_{\phi}(x_{i}^{\textrm{tr}})$ and $g_{\varphi}(\bm p_{j}^{\textrm{s}})$. Then, the concatenated embedding of $f_{\phi}(x_{i}^{\textrm{tr}})$ and $g_{\varphi}(\bm p_{j}^{\textrm{s}})$ is fed into the relation prediction module $g_{\psi}$, which eventually produces a relation score $r_{ij}$ measuring the similarity between $x_{i}^{\textrm{tr}}$ and $\bm p_{j}^{\textrm{s}}$. Thus, we generate $N_{\textrm{s}}$ scores for the relation between the sample $x_{i}^{\textrm{tr}}$ and each seen class prototype by
\begin{align}\label{relation_score}
    \begin{split}
    &r_{ij}=g_{\psi}\left(\left[f_{\phi}^{\text{T}}\left(x_{i}^{\textrm{tr}}\right),g_{\varphi}^{\text{T}}\left(\bm p_{j}^{\textrm{s}}\right)\right]\right), \quad j=1,2,\cdots,N_{\textrm{s}}.     \\
    \end{split}
\end{align}

It is worth noting that we choose concatenation operator, instead of the popularly used product operator (i.e., $f_{\phi}^{\text{T}}(x_{i}^{\textrm{tr}})g_{\varphi}(\bm p_{j}^{\textrm{s}})$) in previous works, to avoid amplifying the small perturbations of input. Actually, the attribute embedding $g_{\varphi}(\bm p_{j}^{\textrm{s}})$ serves as an additional feature map to enhance the discriminability of input image with the original feature map $f_{\phi}(x_{i}^{\textrm{tr}})$. More importantly, concatenation operator can mitigate the domain shift caused by appearance variations of each attribute across all classes. 

\textbf{Relation Based Cross Entropy Loss.} In our ATZSL framework, the probability of a sample $x_{i}^{\textrm{tr}}$ belonging to the $j$-th seen class (i.e., $c_{j}^{\textrm{s}}$) can be measured by the relation between $x_{i}^{\textrm{tr}}$ and $\bm p_{j}^{\textrm{s}}$, i.e.,
\begin{align}\label{probability}
    \begin{split}
    &p\left(c_{j}^{\textrm{s}}|x_{i}^{\textrm{tr}}\right)= \frac{\exp\left(r_{ij}/\gamma\right) }{\sum_{k=1}^{N_{\text{s}}}\exp\left(r_{ik}/\gamma\right)},\\
    \end{split}
\end{align}
where $\gamma$ is the temperature to mitigate overfitting~\cite{hinton2015distilling}, and $\gamma=1$ is a common option. In particular, $\gamma$ “softens” the softmax (raises the output entropy) with $\gamma>1$. As $\gamma \rightarrow \infty$, 
$p(c_{j}^{\textrm{s}}|x_{i}^{\textrm{tr}})\rightarrow
1/N_{\text{s}}$, which leads to maximum uncertainty. As $\gamma \rightarrow 0$, the probability collapses to a point mass (i.e. 
$p(c_{j}^{\textrm{s}}|x_{i}^{\textrm{tr}})=1$). 
Since $\gamma$ does not change the maximum of the softmax function, the class prediction remains unchanged if $\gamma \neq 1$ is applied after convergence.
Plugging the probability in Eq.~(\ref{probability}) into cross-entropy loss over the training sample $x_{i}^{\textrm{tr}}$ from the seen classes $\mathcal{S}$ yields
\begin{align}\label{loss}
    \begin{split}
    &\mathcal{L}\left(\phi,\varphi,\psi;x_{i}^{\textrm{tr}},y_{i}^{\textrm{tr}},{\mathcal{P}}^{\text{s}}\right)  = \sum_{j=1}^{N_{\text{s}}}q_{ij}^{\textrm{tr}}\log p\left(c_{j}^{\textrm{s}}|x_{i}^{\textrm{tr}}\right),\\
    \end{split}
\end{align}
where ${\mathcal{P}}^{\text{s}}=\{\bm p_{i}^{\textrm{s}}\}_{i=1}^{N_{\text{s}}}$.
$q_{ij}^{\textrm{tr}}=1$ if $y_{i}^{\textrm{tr}}=c_{j}^{\textrm{s}}$ and $0$ otherwise. 

\textbf{ATZSL Formulation.} To improve the robustness of our ZSL model (corresponding to the first goal in Section~\ref{Setup}), we further inject adversarial examples into the training set during model training~\cite{goodfellow2014explaining, kurakin2016adversarial}. Specifically, when a clean image $x_{i}^{\textrm{tr}}$ comes to the ZSL model, the attacker is allowed to perturb the image into $x_{i}^{{\textrm{tr}}'}$ with a bounded magnitude. Then based on the loss function in Eq.~(\ref{loss}), we define the overall optimization formulation of the proposed ATZSL as
\begin{align}\label{ATZSL}
    \begin{split}
    \min_{\phi,\varphi,\psi}  &\sum_{(x_{i}^{\textrm{tr}},y_{i}^{\textrm{tr}})\in{\mathcal{D}}^{\textrm{tr}}}\{\alpha \mathcal{L}(\phi,\varphi,\psi;x_{i}^{\textrm{tr}},y_{i}^{\textrm{tr}},{\mathcal{P}}^{\text{s}})\\
    &+(1-\alpha)\max_{x_{i}^{{\textrm{tr}}'}\in B_{\infty}^{\rho}(x_{i}^{\textrm{tr}})} \mathcal{L}(\phi,\varphi,\psi;x_{i}^{{\textrm{tr}}'},y_{i}^{\textrm{tr}},{\mathcal{P}}^{\text{s}})\},\\
    \end{split}
\end{align}
where $B_{\infty}^{\rho}(x_{i}^{\textrm{tr}}):=\{x_{i}^{{\textrm{tr}}'}:\| x_{i}^{{\textrm{tr}}'}-x_{i}^{\textrm{tr}} \|_{{\infty}} \leq \rho\}$, and $\rho\geq 0$ denotes the predefined bound for the attack magnitude. 
The second term in Eq.~(\ref{ATZSL}) aims to produce the adversarial samples by maximally deteriorating the ZSL model performance. 
$\alpha \in [0,1]$ balances the effects of clean and adversarial samples on the model training. 

During the test phase, the predicted class $y_{i}^{\textrm{te}}$ of sample $x_{i}^{\textrm{te}}$ is given by
$y_{i}^{\textrm{te}} = \arg \max_{c_{j}^{\textrm{u}}\in\mathcal{U}} p(c_{j}^{\textrm{u}}|x_{i}^{\textrm{te}})$ in the \underline{standard ZSL} case, and 
$y_{i}^{\textrm{te}} = \arg \max_{c_{j}\in\{\mathcal{S} \cup  \mathcal{U}\}} p(c_{j}|x_{i}^{\textrm{te}})$ in the \underline{generalized ZSL} case. The latter is more practical for real applications since prediction is made over both seen and unseen classes~\cite{zhu2019generalized}.
    
\subsection{ATZSL: Algorithm}\label{ModelOptimization}
It is not trivial to solve the optimization problem in Eq.~(\ref{ATZSL}), since the last term in the objective function is a maximization optimization problem. Considering the last term as a new function, i.e., 
$\mathcal{L}^{\textrm{adv}}(\phi,\varphi,\psi;x_{i}^{\textrm{tr}},y_{i}^{\textrm{tr}},{\mathcal{P}}^{\text{s}})=\max_{x_{i}^{{\textrm{tr}}'}\in B_{\infty}^{\rho}(x_{i}^{\textrm{tr}})} \mathcal{L}(\phi,\varphi,\psi;x_{i}^{{\textrm{tr}}'},y_{i}^{\textrm{tr}},{\mathcal{P}}^{\text{s}})$, then we can solve Eq.~(\ref{ATZSL}) as a standard minimization problem using Adam~\cite{kingma2014adam}. The main bottleneck is to determine the gradient of $\mathcal{L}^{\textrm{adv}}(\cdot)$. We refer to the following theorem~\cite{danskin2012theory}:

\begin{theorem}
    (Danskin's theorem) Let $g(\bm{x},\bm{y})$ ($\bm{x}\in \mathbb{R}^d, \bm{y}\in \Omega$ is a compact set) be a differential function in terms of two vector variables $\bm{x},\bm{y}$. Assume for every $\bm{y}\in \Omega$, $g(\bm{x},\bm{y})$ is convex in $\bm{x}$. Further assume $G(\bm{x}) = \max_{\bm{y}\in \Omega} g(\bm{x},\bm{y})$ is also a differential function in terms of $\bm{x}$, and for any $\bm{x}$, the optimal $\bm{y}^{\ast}(\bm{x}) = \arg \max_{\bm{y}\in \Omega} g(\bm{x},\bm{y})$ is unique in $\Omega$. Then we have the following result: $\frac{\mathrm{d}G(\bm{x})}{\mathrm{d}\bm{x}} = \nabla_{\bm{x}}g(\bm{x},\bm{y}^{\ast}(\bm{x}))$.
\end{theorem}
Based on this theorem, instead of computing the gradient of $\mathcal{L}^{\textrm{adv}}(\cdot)$ directly, we can first find the worst adversarial image by $x_{i}^{{\textrm{tr}}'\ast}=\arg \max_{x_{i}^{{\textrm{tr}}'}\in B_{\infty}^{\rho}(x_{i}^{\textrm{tr}})} \mathcal{L}(\phi,\varphi,\psi;x_{i}^{{\textrm{tr}}'},y_{i}^{\textrm{tr}},{\mathcal{P}}^{\text{s}})$, and then we have $\nabla_{\phi}\mathcal{L}^{\textrm{adv}}(\cdot)=\nabla_{\phi}\mathcal{L}(x_{i}^{{\textrm{tr}}'\ast},y_{i}^{\textrm{tr}},{\mathcal{P}}^{\text{s}};\phi,\varphi,\psi)$ (the same for $\varphi$ and $\psi$).
Thus, the last roadblock is how to find $x_{i}^{{\textrm{tr}}'\ast}$. Inspired by the generation method of adversarial examples in IFGSM attack~\cite{kurakin2016adversarial}, we obtain $x_{i}^{{\textrm{tr}}'\ast}$ using the sign stochastic gradient descent algorithm, where the update formula for the $i$-th step of $N$ iterations is:
$x_{i}^{{\textrm{tr}}'}\leftarrow \textrm{Proj}_{B_{\infty}^{\rho}(x_{i}^{\textrm{tr}})} \{x_{i}^{{\textrm{tr}}'}+\epsilon \textrm{sign}(\nabla_{x_{i}^{\textrm{tr}}}\mathcal{L}(x_{i}^{{\textrm{tr}}'},y_{i}^{\textrm{tr}},{\mathcal{P}}^{\text{s}};\phi,\varphi,\psi))\}$, and $\epsilon \leftarrow 1.25 \rho / N$. A recent study~\cite{bernstein2018signsgd} has validated the convergence of sign (stochastic) gradient descent algorithm. Algorithm~\ref{alg:algorithm} shows the detailed steps of our ATZSL algorithm against image attacks.

\subsection{Extension}
Besides providing improved robustness for ZSL in the visual space, ATZSL can be easily extended to deal with the attacks in semantic space, corresponding to the second goal in Section~\ref{Setup}.
Specifically, one unseen class prototype provided in ZSL can be considered directly analogous to one testing image, and thus can be attacked unavoidably like testing images. To deal with such adversarial attribute attacks effectively, we need to modify our framework in Eq.~(\ref{ATZSL}) as follows:
\begin{align}
  \begin{split}\label{atta}
   \min_{\phi,\varphi,\psi}  &\sum_{(x_{i}^{\textrm{tr}},y_{i}^{\textrm{tr}})\in{\mathcal{D}}^{\textrm{tr}}}\{\alpha \mathcal{L}(\phi,\varphi,\psi;x_{i}^{\textrm{tr}},y_{i}^{\textrm{tr}},{\mathcal{P}}^{\text{s}})\\
    &+(1-\alpha)\max_{\bm p_{i}^{{\textrm{s}}'}\in B_{2}^{\rho}(\bm p_{i}^{\textrm{s}}),\forall i } \mathcal{L}(\phi,\varphi,\psi;x_{i}^{\textrm{tr}},y_{i}^{\textrm{tr}},{\mathcal{P}}^{\text{s}'})\},\\
    \end{split}
\end{align}
where $B_{2}^{\rho}(\bm p_{i}^{\textrm{s}}):=\{\bm p_{i}^{{\textrm{s}}'}:\| \bm p_{i}^{{\textrm{s}}'}-\bm p_{i}^{\textrm{s}} \|_{2} \leq \rho\}$, and ${\mathcal{P}}^{\text{s}'}=\{\bm p_{i}^{{\textrm{s}}'}\}_{i=1}^{N_{\textrm{s}}}$. 
The second term in Eq.~(\ref{atta}) aims to produce the adversarial class prototypes by maximally deteriorating the ZSL model performance. 
It is worth noting that for attack bound constraint, different from the $\ell_{\infty}$ attack in the visual space, $\ell_{2}$ attack is more reasonable in this scenario. This is because $\rho$ denotes the max change of a pixel for image attacks, while the max total change of all attributes for attribute attacks.
The detailed algorithm for Eq.~(\ref{atta}) is provided in Algorithm~\ref{alg:algorithm_att}. 

\begin{algorithm}[tb]
    \caption{ATZSL for image attacks}
    \label{alg:algorithm}
    \KwIn{Dataset $\mathcal{D}^{\textrm{tr}}$, prototypes $\mathcal{P}^{\textrm{s}}$, stepsize sequence ${\{\lambda_t>0\}}_{t=0}^{T-1}$, update steps $N$ and $T$, temperature $\gamma$, and weight $\alpha$.}
    Initialize $\phi,\varphi,\psi$\;
   \For{$t = 0$ to $T-1$}{
        Sample $(x_{i}^{\textrm{tr}},y_{i}^{\textrm{tr}})$ from $\mathcal{D}^{\textrm{tr}}$\;
        Draw $\rho \in [0,4]$ from a truncated normal distribution $\mathcal{N}(0,2)$\;
        $\epsilon = 1.25 \rho / N$\;
        $x_{i}^{{\textrm{tr}}'}\leftarrow x_{i}^{\textrm{tr}}$\;
        \For{$i = 0$ to $N-1$}{
           $\Delta = \epsilon \textrm{sign}(\nabla_{x_{i}^{\textrm{tr}}}\mathcal{L}(x_{i}^{{\textrm{tr}}'},y_{i}^{\textrm{tr}},{\mathcal{P}}^{\text{s}};\phi,\varphi,\psi))$\;
           $x_{i}^{{\textrm{tr}}'}\leftarrow \textrm{Proj}_{B_{\infty}^{\rho}(x_{i}^{\textrm{tr}})} \{x_{i}^{{\textrm{tr}}'}+\Delta\}$\;
        }
        $x_{i}^{{\textrm{tr}}'\ast} \leftarrow x_{i}^{{\textrm{tr}}'}$\;
        Update $\phi,\varphi,\psi$ using Adam\ with $\lambda_t$;
    }
        \KwOut{Model parameters $\phi,\varphi,\psi$ }
\end{algorithm}

\begin{algorithm}[htb]
    \caption{ATZSL for attribute attacks}
    \label{alg:algorithm_att}
    \KwIn{Dataset $\mathcal{D}^{\textrm{tr}}$, prototypes $\mathcal{P}^{\textrm{s}}$, stepsize sequence ${\{\lambda_t>0\}}_{t=0}^{T-1}$, update steps $N$ and $T$, temperature $\gamma$, and weight $\alpha$.}
    Initialize $\phi,\varphi,\psi$\;
   \For{$t = 0$ to $T-1$}{
        Sample $(x_{i}^{\textrm{tr}},y_{i}^{\textrm{tr}})$ from        $\mathcal{D}^{\textrm{tr}}$\;
        \For{$i = 0$ to $N-1$}{
           Sample $\bm {p}^{\textrm{s}}$ from  ${\mathcal{P}}^{\text{s}}$\;
           Draw $\rho \in [0,54]$ from a truncated normal distribution $N(\mu =0,\sigma =27)$\;
           $\epsilon \leftarrow \sqrt{\rho^{2}/\left | \bm p_{i}^{\textrm{s}} \right |}$\;
           ${\mathcal{P}}^{\text{s}'}\leftarrow {\mathcal{P}}^{\text{s}}$\;
           $\Delta = \epsilon \textrm{sign}(\nabla_{\bm p_{i}^{\textrm{s}}}\mathcal{L}(x_{i}^{{\textrm{tr}}},y_{i}^{\textrm{tr}},{\mathcal{P}}^{\text{s}'};\phi,\varphi,\psi))$\;
           $ \bm {p}^{\textrm{s}'}\leftarrow \textrm{Proj}_{B_{2}^{\rho}(\bm p_{i}^{\textrm{s}})} \{\bm p_{i}^{{\textrm{s}}'}+\Delta\}$\;
        }
        ${\mathcal{P}}^{\text{s}'\ast} \leftarrow {\mathcal{P}}^{\text{s}'}$\;
        Update $\phi,\varphi,\psi$ using Adam\ with $\lambda_t$;
    }
        \KwOut{Model parameters $\phi,\varphi,\psi$ }
\end{algorithm}

\section{Evaluation Setup and Metrics}
In this section, we first provide the evaluation protocols, e.g. dataset splits and evaluation metrics, and then detail our experimental implementation.

\subsection{Datasets and Protocols} 
Among the most widely used datasets for ZSL, we select one coarse-grained dataset (Animals with Attributes2 (AWA2)\footnote{\url{http://cvml.ist.ac.at/AwA2/}}~\cite{xian2018zero}), and one fine-grained dataset (CUB-200-2011 Birds (CUB)\footnote{\url{http://www.vision.caltech.edu/visipedia/CUB-200-2011.html}}~\cite{wah2011caltech}). Specifically, AWA2 has 85 attributes, containing 37,322 images from 50 different animal classes. On average, each class includes 746 images where the least populated class, i.e. mole, has 100 and the most populated class, i.e. horse has 1645 examples. CUB has a large number of classes and attributes, containing 11,788 images from 200 different types of birds annotated with 312 attributes. 

We adopt the novel rigorous protocol\footnote{\url{http://www.mpi-inf.mpg.de/zsl-benchmark}} proposed in~\cite{xian2018zero}, insuring that none of the unseen classes appear in ImageNet 1K, since ImageNet 1K is used to pre-train the Resnet model. Otherwise the zero-shot rule would be violated. Specifically, AWA2 and CUB datasets consist of 27 and 100 training (seen) classes respectively while 10 and 50 test (unseen) classes. The hyperparameter search is performed on a disjoint set of validation set of 13 and 50 classes respectively. 
In particular, this protocol involves two settings: \emph{Standard ZSL} and \emph{Generalized ZSL}. 
The latter emerges recently under which the test set contains data samples from both seen and unseen classes. This setting is thus clearly more reflective of real-world application scenarios. By contrast, the test set in standard ZSL only contains data samples from the unseen classes.

\subsection{Evaluation Metrics} 
At test phase of ZSL, we use the unified evaluation protocol proposed in~\cite{xian2018zero}. 
Specifically, under the standard ZSL setting, we adopt the average per-class top-1 accuracy ($\textrm{T1}$) in the following way:
\begin{align}
    acc_{\mathcal{U}}=\frac{1}{N_{\textrm{u}}} \sum_{c_{i}^{\textrm{u}} \in \mathcal{U}} \frac{\# {\rm correct~predictions~in}~c_{i}^{\textrm{u}}}{\# {\rm samples~in}~c_{i}^{\textrm{u}}}.\nonumber
\end{align}
Under the generalized ZSL setting, we compute the harmonic mean ($\textrm{HM}$) of $acc_{\mathcal{S}}$ and $acc_{\mathcal{U}}$ to favor high accuracies on both seen and unseen classes as follows:
\begin{align*}
H=\frac{2*acc_{\mathcal{S}}*acc_{\mathcal{U}}}{acc_{\mathcal{S}}+acc_{\mathcal{U}}},\nonumber
\end{align*}
where $acc_{\mathcal{S}}$ and $acc_{\mathcal{U}}$ are the average per-class top-1 accuracies of recognizing the testing samples from the seen and the unseen classes, respectively, and
\begin{align}
    acc_{\mathcal{S}}=\frac{1}{N_{\textrm{s}}} \sum_{c_{i}^{\textrm{s}} \in \mathcal{S}} \frac{\# {\rm correct~predictions~in}~c_{i}^{\textrm{s}}}{\# {\rm samples~in}~c_{i}^{\textrm{s}}}.\nonumber
\end{align}
Finally, to evaluate the overall performance (i.e., trade-off on clean and adversarial data) under the standard ZSL setting, we further compute the harmonic mean (\underline{$\textrm{H}_{\textrm{T1}}$}) of $\textrm{T1}_{\textrm{cle}}$ and $\textrm{T1}_{\textrm{adv}}$, where $\textrm{T1}_{\textrm{cle}}$ and $\textrm{T1}_{\textrm{adv}}$ represent the $\textrm{T1}$ metric in the clean and the adversarial scenarios, respectively.
Similarly, let $\textrm{HM}_{\textrm{cle}}$ and $\textrm{HM}_{\textrm{adv}}$ denote the $\textrm{HM}$ metric in the clean and adversarial scenarios, respectively. The harmonic mean (\underline{$\textrm{H}_{\textrm{HM}}$}) of $\textrm{HM}_{\textrm{cle}}$ and $\textrm{HM}_{\textrm{adv}}$ is used to evaluate the overall performance under the generalized ZSL setting.

\subsection{Implementation Details} 
By tuning on the validation set, we set $\alpha=0.5$, $\gamma=1$, $N=3$, and train ATZSL for 400 epochs on the two datasets with weight decay $10^{-5}$. The learning rate $\lambda_{t}$ is initialized to $10^{-5}$ with Adam and then annealed by 10\% every 10 epochs. Specifically, 
i) For the feature extraction module, before being fed into the pre-trained Resnet34 network\footnote{\url{https://github.com/pytorch/vision/blob/master/torchvision/models/resnet.py}}, a color image is first normalized with the means $\{0.485,0.456,0.406\}$ and standard deviations $\{0.229,0.224,0.225\}$ with respect to each channel. 
ii) For the attribute embedding module, the continuous attributes\footnote{Continuous attributes are also provided by datasets, as binary attributes have been shown to be weaker than continuous attributes.} between values 0 and 100 are fed into the MLP network. The size of hidden layer (Fig.~\ref{fig:Framework}) is set to 300 and 400 for AWA2 and CUB respectively, and the output size is set to the same dimension (512) as the image embedding for both datasets. Additionally, we add weight decay ($\ell_{2}$ regularisation), and ReLU non-linearity for two fully-connected layers.
iii) For the relation prediction module, the image and prototype embeddings are concatenated before being fed into MLPs with hidden layer size 400 for both datasets. The first fully-connected layer is also with ReLU.
We add weight decay ($\ell_{2}$ regularisation) in the attribute embedding module as there is a hubness problem in cross-modal mapping for ZSL which can be best solved by mapping the semantic feature vector to the visual feature space with regularisation. 

\textbf{Compared Methods.}
We choose to compare with several competitive and representative ZSL methods that have achieved the state-of-the-art results recently, including DEM\footnote{\url{https://github.com/lzrobots/DeepEmbeddingModel_ZSL}}~\cite{zhang2017learning}, FGN\footnote{\url{http://datasets.d2.mpi-inf.mpg.de/xian/cvpr18xian.zip}}~\cite{xian2018feature}, MC-ZSL\footnote{\url{https://github.com/rfelixmg/frwgan-eccv18}}~\cite{felix2018multi}, and LDF\footnote{\url{https://github.com/shaoniangu/Zero_shot_learning_using_LDF_tensorflow}}~\cite{li2018discriminative}. 
Specifically, DEM~\cite{zhang2017learning} and LDF~\cite{li2018discriminative} perform better in the second and last group of ZSL methods (described in Section~\ref{Related work}), respectively. FGN~\cite{xian2018feature} and MC-ZSL~\cite{felix2018multi} are Generative Adversarial Network (GAN) based ZSL models.
In addition, to validate the effectiveness of our proposed adversarial training, we evaluate our ATZSL without adversarial samples, i.e., $\alpha=1$, dubbed \textbf{Baseline}. 

During the test phase of each ZSL method, besides using clean data (i.e., ``No attack''), we evaluate the model robustness using five widely used white-box attacks (including FGSM~\cite{goodfellow2014explaining}, IFGSM~\cite{kurakin2016adversarial}, DeepFool~\cite{moosavi2016deepfool}, CW~\cite{carlini2017towards}, and WRM~\cite{sinha2017certifying}), and one black-box attack (ZOO~\cite{chen2017zoo}). 
It is worth noting that these attack methods have been standardized in adversarial learning. 
IFGSM~\cite{kurakin2016adversarial} is a straightforward extension of FGSM~\cite{goodfellow2014explaining} by applying it multiple times with small step size. 
Compared with FGSM~\cite{goodfellow2014explaining}, DeepFool~\cite{moosavi2016deepfool} can generate adversarial perturbations with lower norm.
CW~\cite{carlini2017towards} introduces three new attacks for the $\ell_{0}$, $\ell_{2}$, and $\ell_{\infty}$ distance metrics, and in our experiments, we adopt $\ell_{2}$ untargeted attacks.
WRM~\cite{sinha2017certifying} additionally considers a Lagrangian penalty formulation of perturbation of the underlying data distribution in a Wasserstein ball to achieve worst-case perturbations of training data.
ZOO~\cite{chen2017zoo} is a zeroth order optimization based attack to directly estimate the gradients of the targeted model for generating adversarial examples, and we also adopt the widely used $\ell_{2}$ untargeted attacks to our model.

\begin{figure*}[!tbp]
\centering
\subfigure[Image attack on AWA2]{\label{image_szsl_awa}
\begin{minipage}[t]{0.25\linewidth}
\centering
\includegraphics[width=1.75in]{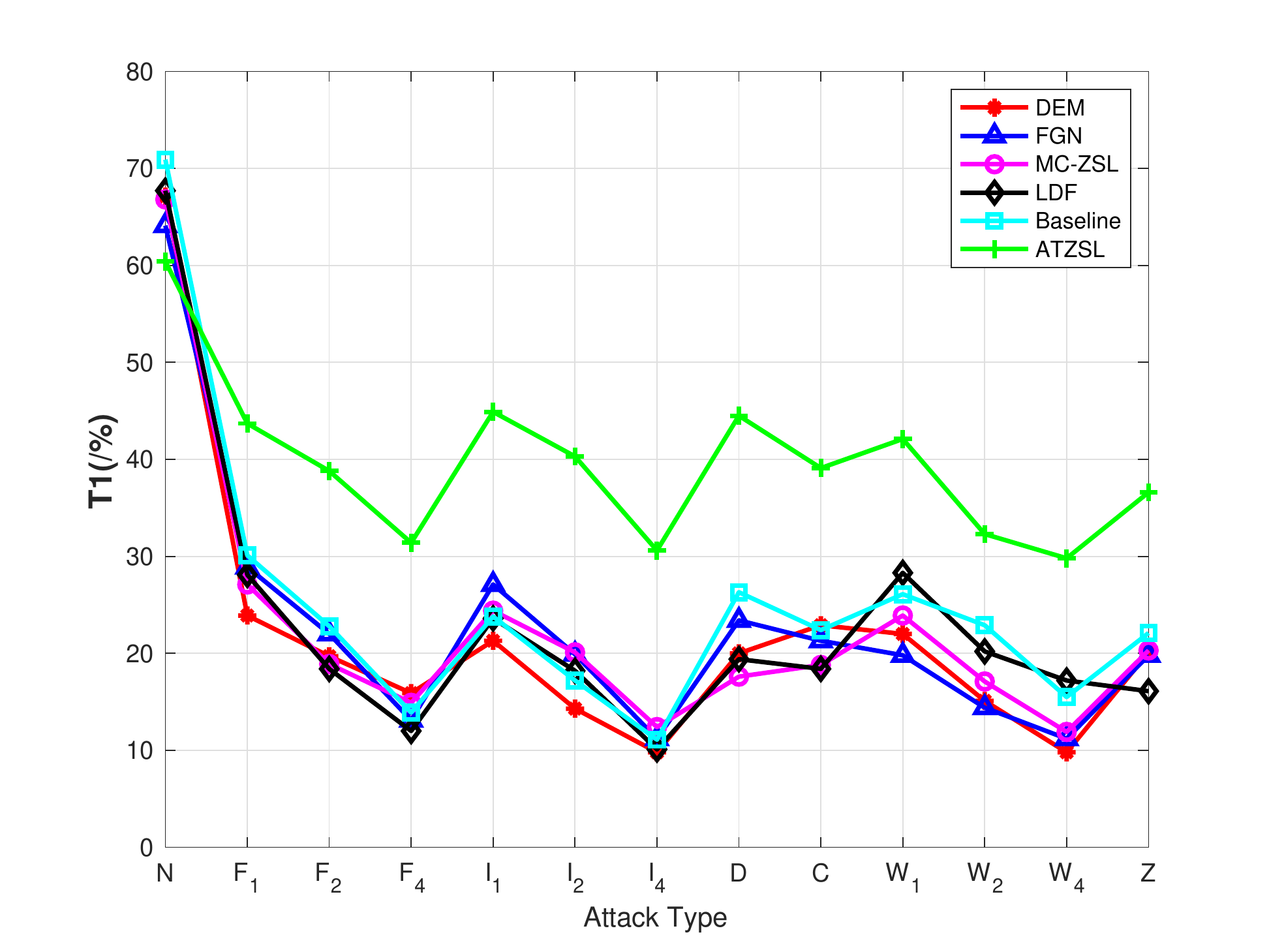}
\end{minipage}%
}%
\subfigure[Attribute attack on AWA2]{\label{att_szsl_awa}
\begin{minipage}[t]{0.25\linewidth}
\centering
\includegraphics[width=1.75in]{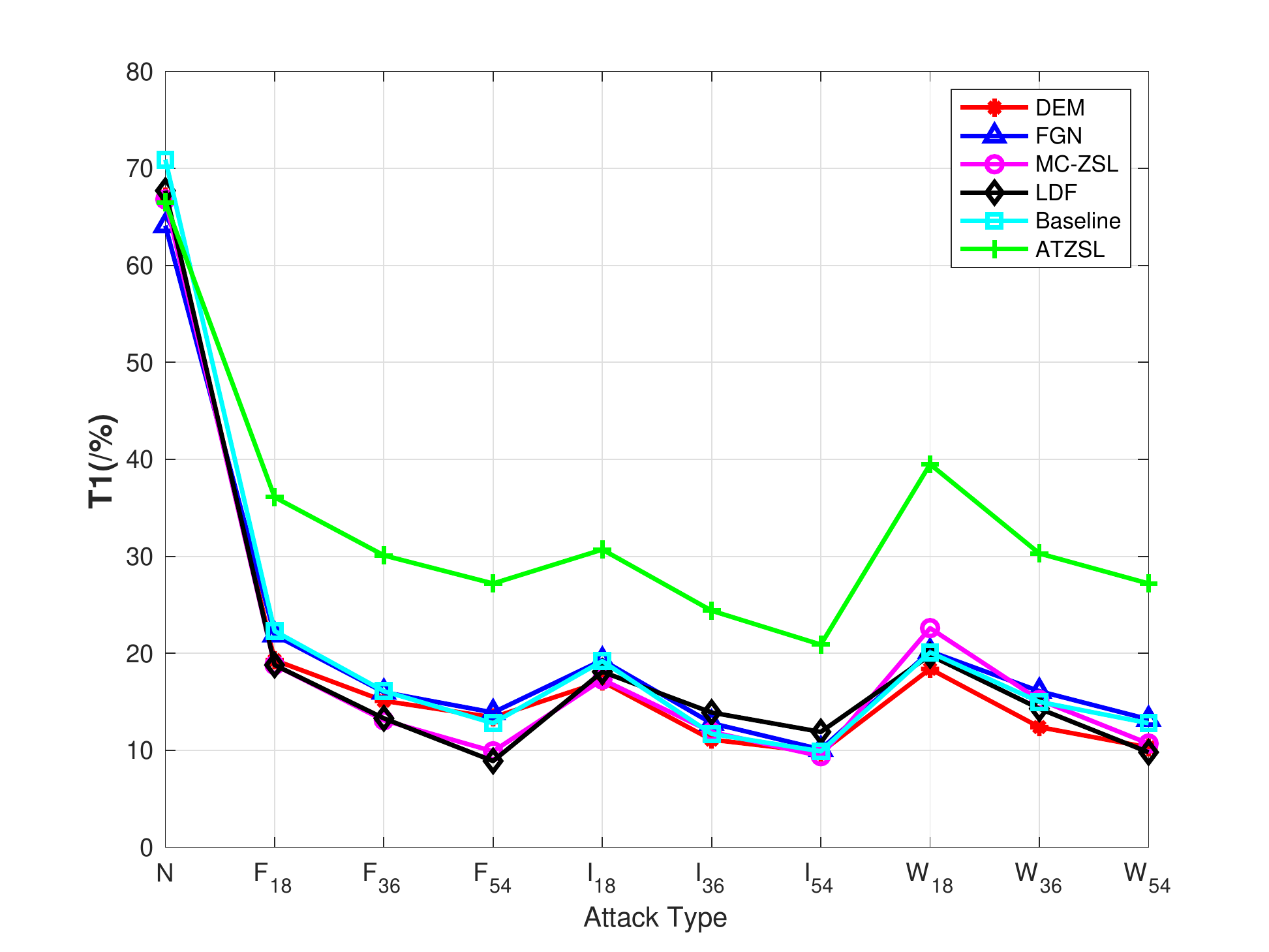}
\end{minipage}%
}%
\subfigure[Image attack on CUB]{\label{image_szsl_cub}
\begin{minipage}[t]{0.25\linewidth}
\centering
\includegraphics[width=1.75in]{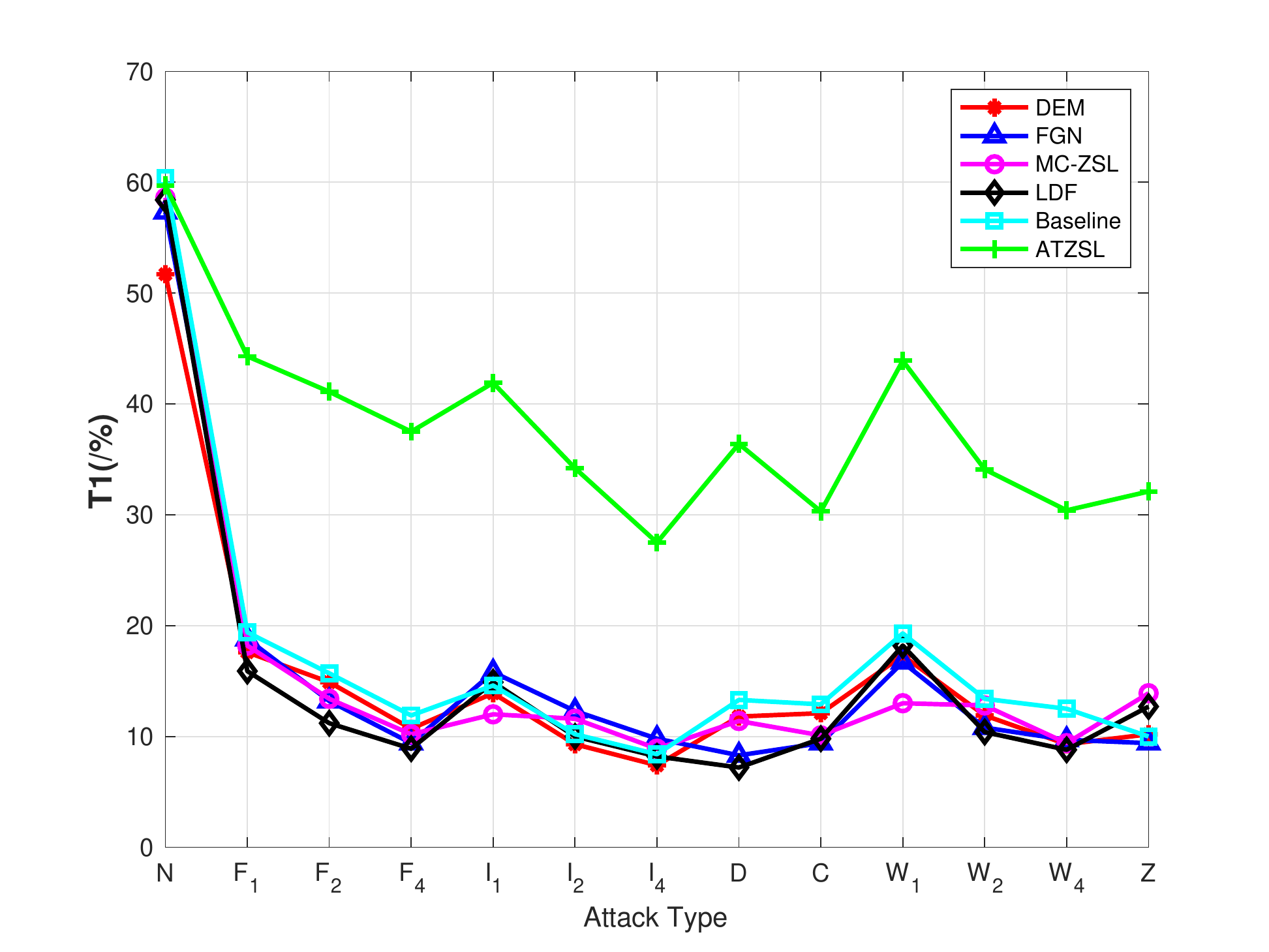}
\end{minipage}
}%
\subfigure[Attribute attack on CUB]{\label{att_szsl_cub}
\begin{minipage}[t]{0.25\linewidth}
\centering
\includegraphics[width=1.75in]{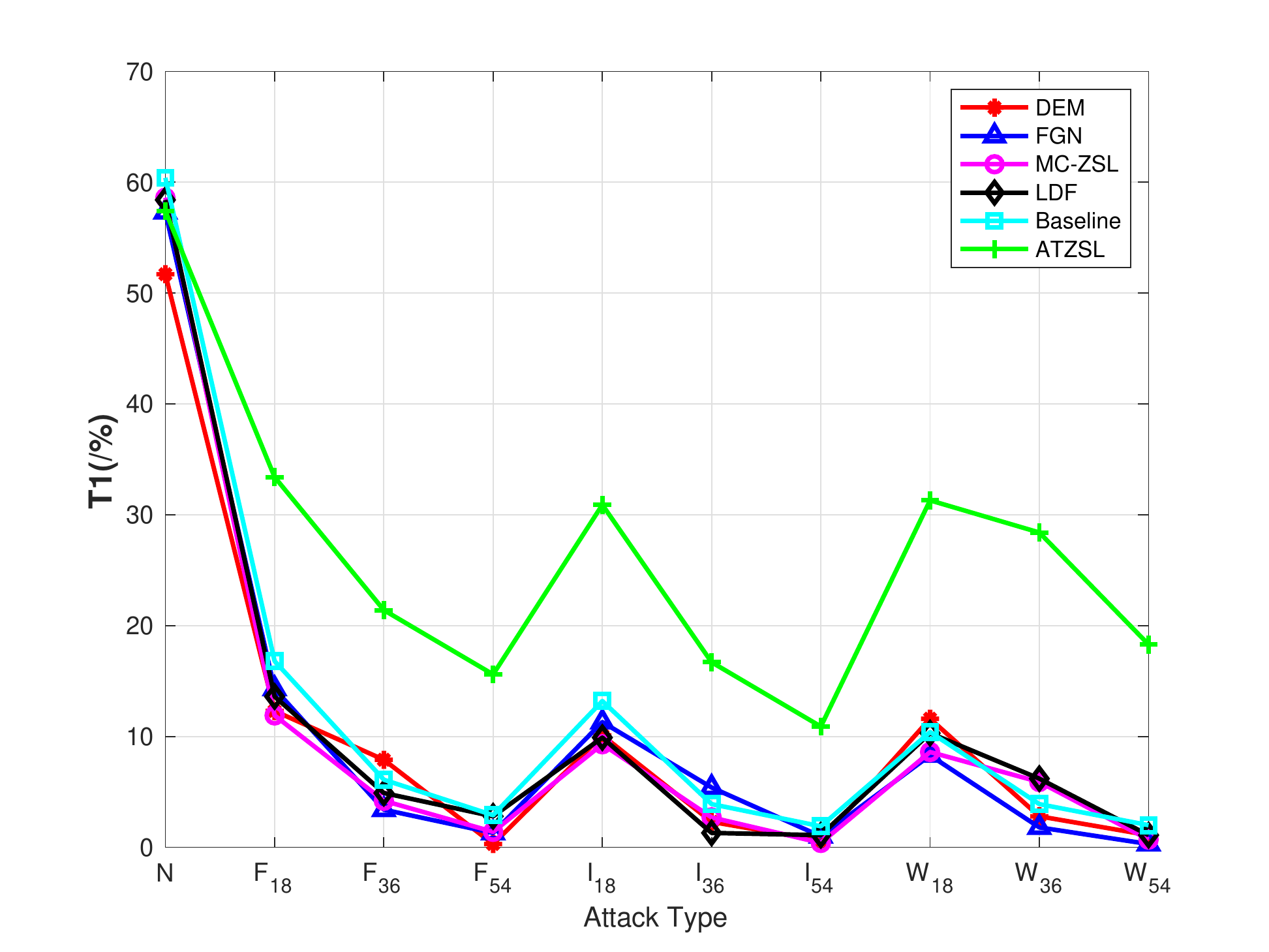}
\end{minipage}
}%
\centering
\caption{Comparison (T1:\%) of several ZSL methods on two datasets under the \textbf{standard} ZSL setting in various attack scenarios. For horizontal axis, 'N', 'F', 'I', 'D', 'C', 'W', and 'Z' represent No attack, FGSM, IFGSM, DeepFool, CW, WRM, and ZOO attack methods, respectively. The numbers with respect to 'F', 'I', and 'W' represent the attack magnitudes.}
\label{fig:ZSL}
\end{figure*} 

\begin{table*}[!t]
\caption{Comparative Results ($\textrm{H}_{\textrm{T1}}$:\%) of Several ZSL Methods on \textbf{AWA2} under the \textbf{Standard} ZSL Setting with Various Attacks in the \textbf{Visual} Space. 
'-' Represents that the Attack Magnitude Is Not Required for That Attack Method.
The Number in Parentheses Measures the Improvement Obtained by Our ATZSL over the Strongest Competitor in Terms of $\textrm{H}_{\textrm{T1}}$
 } \label{tab:AWA2_SZSL_Visual}
  \vspace{-0.3cm}
 \begin{center}
  \begin{tabular}{c|c|cccccc}
    \hline \hline
    \textbf{Attack type} &\textbf{Magnitude} & \textbf{DEM}~\cite{zhang2017learning}& \textbf{FGN}~\cite{xian2018feature} & \textbf{MC-ZSL}~\cite{felix2018multi} & \textbf{LDF}~\cite{li2018discriminative} & \textbf{Baseline}& \textbf{ATZSL}(Ours)\\
    \hline
    \multirow{3}*{\textbf{FGSM~\cite{goodfellow2014explaining}}} & $\rho=1$ &35.2&39.8&	38.6&	39.7&42.3&\textbf{50.7}(8.4)\\ 
    & $\rho=2$ & 30.5&	32.8&	29.5&	28.9&	34.5&	\textbf{47.2}(12.7) \\
    & $\rho=4$ & 25.7&	21.8&	24.4&	20.4&	23.2&	\textbf{41.3}(15.6)\\
    \hline
    \multirow{3}*{\textbf{IFGSM~\cite{kurakin2016adversarial}}} & $\rho=1$ &  32.3&	38.1&	35.7&	35.0&	35.6&	\textbf{51.5}(13.4)\\
    & $\rho=2$ & 23.6&	30.4&	30.9&	28.7&	27.7&	\textbf{48.3}(17.4) \\
    & $\rho=4$ &  17.1&	19.1&	20.9&	17.6&	19.2&	\textbf{40.6}(19.7) \\
    \hline
    \textbf{DeepFool~\cite{moosavi2016deepfool}} &-&30.8&	34.3&	27.9&	30.2&	38.4&\textbf{51.2}(12.8) \\
    \hline
    \textbf{CW~\cite{carlini2017towards}}&-  & 34.1	&32.0&	29.3&	28.9&	34.0&	\textbf{47.5}(13.4)  \\
    \hline
    \multirow{3}*{\textbf{WRM~\cite{sinha2017certifying}}} & $\rho=1$ & 33.1&30.3&35.2&39.9&	38.2&	\textbf{49.6}(9.7) \\
    & $\rho=2$ & 24.7&	23.5&	27.2&	31.1&	34.6&	\textbf{42.1}(7.5) \\
    & $\rho=4$ &  17.1&	19.1&	20.2&	27.4&	25.4&	\textbf{39.9}(12.5) \\
    \hline
    \textbf{ZOO~\cite{chen2017zoo}} & -&  30.9&	30.3&	31.1&	26.0&	33.7&	\textbf{45.6}(11.9) \\
    \hline\hline
    \end{tabular}
 \end{center}
\end{table*}

\textbf{Attack Settings.} i) During model training of ATZSL, we employ the IFGSM attacker~\cite{kurakin2016adversarial} to find adversarial examples. Specifically, for attacks in the visual space, the attack magnitude $\rho$ in each epoch is drawn from a truncated normal distribution defined in interval $[0,4]$ with $\mathcal{N}(0,2)$ for both datasets. Due to different dimensions of attributes, for attacks in the semantic space, $\rho \in [0,27]$ with underlying normal distribution $\mathcal{N}(0,14)$ for AWA2, and $\rho \in [0,54]$ with underlying normal distribution $\mathcal{N}(0,27)$ for CUB.
ii) During the test phase, we adopt five widely used white-box attacks (including FGSM~\cite{goodfellow2014explaining}, IFGSM~\cite{kurakin2016adversarial}, DeepFool~\cite{moosavi2016deepfool}, CW~\cite{carlini2017towards}, and WRM~\cite{sinha2017certifying}), and one black-box attack (ZOO~\cite{chen2017zoo}). Specifically, for FGSM~\cite{goodfellow2014explaining}, IFGSM~\cite{kurakin2016adversarial}, and WRM~\cite{sinha2017certifying} attacks, we fix the image perturbation magnitude as $\rho\in \{1,2,4\}$ for both datasets in the visual space. 
Specially, $\rho$ has the same range for any datasets since each pixel is in $[0,255]$. Generally, $\rho \leq  8$ for imperceptibility. 
While in the semantic space, we fix the attribute perturbation magnitude as $\rho\in \{9,18,27\}$ and $\rho\in \{18,36,54\}$ for AWA2 and CUB respectively.
That is, the attribute attack magnitude on CUB is larger than that on AWA2, since CUB has 312 attributes while AWA2 has only 85.
Additionally, we set the number of attack iterations as $9$ for IFGSM~\cite{kurakin2016adversarial}, DeepFool~\cite{moosavi2016deepfool}, CW~\cite{carlini2017towards}, and WRM~\cite{sinha2017certifying} attacks. 

\begin{table*}[!t]
  \caption{Comparative Results ($\textrm{H}_{\textrm{T1}}$:\%) of Several ZSL Methods on \textbf{AWA2} under the \textbf{Standard} ZSL with Various Attacks in the \textbf{Semantic} Space.
 The Number in Parentheses Measures the Improvement Obtained by Our ATZSL over the Strongest Competitor in Terms of $\textrm{H}_{\textrm{T1}}$
 } \label{tab:AWA2_SZSL_Semanticl}
 \vspace{-0.3cm}
 \begin{center}
  \begin{tabular}{c|c|cccccc}
    \hline \hline
    \textbf{Attack type} &\textbf{Magnitude} & \textbf{DEM}~\cite{zhang2017learning}& \textbf{FGN}~\cite{xian2018feature} & \textbf{MC-ZSL}~\cite{felix2018multi} & \textbf{LDF}~\cite{li2018discriminative} & \textbf{Baseline}& \textbf{ATZSL}(Ours)\\
    \hline
    \multirow{3}*{\textbf{FGSM~\cite{goodfellow2014explaining}}} & $\rho=9$ &  30.0&	32.6&	29.3&	29.4&	33.9&	\textbf{46.8}(12.9)\\ 
    & $\rho=18$ & 24.7&	25.6&	21.9&	22.2&	26.2&	\textbf{41.4}(15.2) \\
    & $\rho=27$ & 22.3&	22.8&	17.2&	15.7&	21.7&	\textbf{38.6}(15.8)\\
    \hline
    \multirow{3}*{\textbf{IFGSM~\cite{kurakin2016adversarial}}} & $\rho=9$ & 27.3&	29.7&	27.5&	28.6&	30.2&	\textbf{42.0}(11.8) \\
    & $\rho=18$ & 19.0&	21.3&	20.2&	23.1&	20.1&	\textbf{35.7}(12.6)\\
    & $\rho=27$ & 17.1&	17.5&	16.5&	20.2&	17.4&	\textbf{31.8}(11.6)\\
    \hline
    \multirow{3}*{\textbf{WRM~\cite{sinha2017certifying}}} & 
    $\rho=9$ & 28.9	&30.7	&33.8	&30.6	&31.3&	\textbf{49.6}(15.8) \\
    & $\rho=18$ & 20.9	&25.7&	24.8&	23.6&	24.8&	\textbf{41.6}(15.9) \\
    & $\rho=27$ & 17.9&	21.9&	18.4&	17.1&	21.7&	\textbf{38.6}(16.7)\\
    \hline\hline
    \end{tabular}
 \end{center}
\end{table*}

\begin{table*}[!t]
\caption{Comparative Results ($\textrm{H}_{\textrm{T1}}$:\%) of Several ZSL Methods on \textbf{CUB} under the \textbf{Standard} ZSL with Various Attacks in the \textbf{Visual} Space. 
'-' Represents that the Attack Magnitude Is Not Required for That Attack Method.
The Number in Parentheses Measures the Improvement Obtained by Our ATZSL over the Strongest Competitor in Terms of $\textrm{H}_{\textrm{T1}}$
 } \label{tab:CUB_SZSL_Visual}
 \vspace{-0.3cm}
 \begin{center}
  \begin{tabular}{c|c|cccccc}
    \hline \hline
    \textbf{Attack type} &\textbf{Magnitude} & \textbf{DEM}~\cite{zhang2017learning}& \textbf{FGN}~\cite{xian2018feature} & \textbf{MC-ZSL}~\cite{felix2018multi} & \textbf{LDF}~\cite{li2018discriminative} & \textbf{Baseline}& \textbf{ATZSL}(Ours)\\
    \hline
    \multirow{3}*{\textbf{FGSM~\cite{goodfellow2014explaining}}} & $\rho=1$ & 26.3&	28.3&	27.8&	25.0&	29.4&	\textbf{50.9}(21.5)\\ 
    & $\rho=2$ & 23.1&	21.5&	21.8&	18.8&	24.9&	\textbf{48.7}(23.8) \\
    & $\rho=4$ &  17.7&	16.2&	17.4&	15.4&	19.9&	\textbf{46.1}(26.2)\\
    \hline
    \multirow{3}*{\textbf{IFGSM~\cite{kurakin2016adversarial}}} & $\rho=1$ & 21.9&	24.8&	19.9&	23.7&	23.5&	\textbf{49.2}(24.4) \\
    & $\rho=2$ &  15.8&	20.3&	19.4&	17.1&	17.5&	\textbf{43.5}(23.2)\\
    & $\rho=4$ & 12.9&	16.7&	15.5&	14.4&	14.7&	\textbf{37.7}(21.0) \\
    \hline
    \textbf{DeepFool~\cite{moosavi2016deepfool}} &- & 19.2&	14.5&	19.1&12.8&	21.8&	\textbf{45.2}(23.4) \\
    \hline
    \textbf{CW~\cite{carlini2017towards}}&-  & 19.6&	16.2&	17.2&	16.8&	21.3&	\textbf{40.2} (18.9) \\
    \hline
    \multirow{3}*{\textbf{WRM~\cite{sinha2017certifying}}} & $\rho=1$ & 25.9&	25.9&	21.3&	27.8&	29.3&	\textbf{50.6}(21.3) \\
    & $\rho=2$ & 19.3 &	18.2&	21.0&	17.7&	21.9&	\textbf{43.4} (21.5)\\
    & $\rho=4$ & 15.8 & 16.6&	16.2&	15.3&	20.7&	\textbf{40.3}(19.6) \\
    \hline
    \textbf{ZOO~\cite{chen2017zoo}} & -&  17.0&	16.2&	22.5&	20.9&	17.2&	\textbf{41.8}(19.3)\\
    \hline\hline
    \end{tabular}
 \end{center}
\end{table*}

\section{Experimental Results and Analysis}
We first present standard ZSL results on AWA2 and CUB datasets with various attacks in visual and semantic spaces. Then, we provide results for the generalized ZSL setting. Additionally, several ablation experiments are further conducted to demonstrate the effectiveness of our ATZSL. Finally, we also provide complexity analysis of our method to show its efficiency.

\subsection{Standard ZSL} \label{sec:szsl}
To evaluate the effectiveness of our proposed ATZSL under the standard ZSL setting, we first perform recognition task on AWA2 and CUB datasets without any attacks, and then with six typical attack methods, including FGSM~\cite{goodfellow2014explaining}, IFGSM~\cite{kurakin2016adversarial}, DeepFool~\cite{moosavi2016deepfool}, CW~\cite{carlini2017towards}, WRM~\cite{sinha2017certifying}, and ZOO~\cite{chen2017zoo}. 

\textbf{Standard ZSL Results on AWA2.} 
We first evaluate the robustness and transferability of ATZSL in Eq.~(\ref{ATZSL}), where the adversarial images are generated to train our ZSL model. 
Fig.~\ref{image_szsl_awa} shows the compared results on AWA2 dataset in terms of average per-class top-1 accuracy ($\textrm{T1}$) with other five ZSL methods, i.e., DEM~\cite{zhang2017learning}, FGN~\cite{xian2018feature}, MC-ZSL~\cite{felix2018multi}, LDF~\cite{li2018discriminative}, and our \text{Baseline}. 
Specifically, the performance of standard ZSL without any attacks (i.e., $\textrm{T1}_{\textrm{cle}}$) for each method is denoted by `N' in horizontal axis.
And the remainder in horizontal axis denote the performances with six types of attacks (i.e., $\textrm{T1}_{\textrm{adv}}$) in the visual space, where we also set three kinds of attack magnitudes (i.e., $\rho\in\{1,2,4\}$) for FGSM~\cite{goodfellow2014explaining}, IFGSM~\cite{kurakin2016adversarial}, and WRM~\cite{sinha2017certifying} attacks.
To illustrate the trade-off of each method on clean and adversarial samples more straightforwardly, we further provide the harmonic mean ({$\textrm{H}_{\textrm{T1}}$}) of $\textrm{T1}_{\textrm{cle}}$ and $\textrm{T1}_{\textrm{adv}}$ in Table~\ref{tab:AWA2_SZSL_Visual}.
By contrast, it can be observed that our ATZSL obtains remarkably more favorable trade-off of zero-shot recognition performances on \underline{clean and adversarial images} of AWA2, over currently available alternatives. The improvements achieved by our model over the strongest competitor range from 7.5\% to 19.7\%. 
Additionally, Fig.~\ref{image_szsl_awa} shows that image attacks deteriorate the performance of each compared ZSL method under no attack scenario (i.e., $\textrm{T1}_{\textrm{cle}}$). However, our ATZSL shows the most non-obvious degradation in various image attack scenarios. Therefore, ATZSL paves a possible direction to improve robustness of ZSL.

Then, we evaluate the robustness and transferability of ATZSL in Eq.~(\ref{atta}), where the adversarial attributes of seen classes are generated to train our ZSL model. 
Fig.~\ref{att_szsl_awa} shows the compared results on AWA2 in terms of average per-class top-1 accuracy ($\textrm{T1}$) with five ZSL methods.
`N' in horizontal axis still denotes the performance without any attacks (i.e., $\textrm{T1}_{\textrm{cle}}$) for each method, and the remainder represent $\textrm{T1}_{\textrm{adv}}$ under three types of attacks, i.e., FGSM~\cite{goodfellow2014explaining}, IFGSM~\cite{kurakin2016adversarial}, and WRM~\cite{sinha2017certifying} with $\rho\in\{9,18,27\}$.
Correspondingly, Table~\ref{tab:AWA2_SZSL_Semanticl} presents the harmonic mean ({$\textrm{H}_{\textrm{T1}}$}), i.e., the trade-off of each method on clean and adversarial attributes.
As expected, our ATZSL can also obtain more promising trade-off of zero-shot recognition performances on \underline{clean and adversarial attributes} of AWA2 over all competitors, generally more than 11.6\%. Meanwhile, as shown in Fig.~\ref{att_szsl_awa}, our ATZSL can achieve the consistently slightest performance degradation in various attribute attack scenarios compared with no attack case.

\textbf{Standard ZSL Results on CUB.} 
Likewise, under the standard ZSL setting, we further evaluate the performances of ATZSL in Eq.~(\ref{ATZSL}) and Eq.~(\ref{atta}) on CUB dataset. 
First, we train our ATZSL by Eq.~(\ref{ATZSL}), and test on both clean and adversarial images of CUB. 
The compared results in terms of average per-class top-1 accuracy ($\textrm{T1}$) with five representative ZSL methods are plotted in Fig.~\ref{image_szsl_cub}, where `N' in horizontal axis denotes the performance without any attacks ({i.e.} $\textrm{T1}_{\textrm{cle}}$), and the remainder represent $\textrm{T1}_{\textrm{adv}}$ under six types of attacks with three types of magnitudes.
Table~\ref{tab:CUB_SZSL_Visual} additionally presents the harmonic mean ({$\textrm{H}_{\textrm{T1}}$}) of $\textrm{T1}_{\textrm{cle}}$ and $\textrm{T1}_{\textrm{adv}}$, to show the trade-off of each method on clean and adversarial images. We can find that the improvements obtained by our ATZSL over the strongest competitor are averagely around 21.5\%.
Then, we train our ATZSL by Eq.~(\ref{atta}), and test on both clean and adversarial attributes of CUB. Fig.~\ref{att_szsl_cub} shows the average per-class top-1 accuracy ($\textrm{T1}$) of each method under various attack settings, including no attack case ($\textrm{T1}_{\textrm{cle}}$) and three kinds of attacks with three types of magnitudes ($\textrm{T1}_{\textrm{adv}}$). The trade-off of each method on clean and adversarial attributes is quantitatively compared in Table~\ref{tab:CUB_SZSL_Semanticl} with harmonic mean ({$\textrm{H}_{\textrm{T1}}$}) of $\textrm{T1}_{\textrm{cle}}$ and $\textrm{T1}_{\textrm{adv}}$.
In this scenario, our method still can improve the trade-off performance over the strongest competitor by an obvious margin (14.6\%$\sim$26.8\%).
From these results on CUB, we can draw similar conclusions with AWA2. Specially, the improvements obtained by our ATZSL on CUB are more significant than those on AWA2, for both image and attribute attacks. This is possibly due to the fact that the samples in CUB are fine-grained, and thus we can generate more reliable adversarial samples to train a more robust ZSL model.

In summary, the four tables demonstrate that our ATZSL can achieve obviously more competitive trade-off of standard zero-shot recognition performances on clean and adversarial data. More importantly, such obvious improvements have not sacrificed the $\textrm{T1}$ performance of our ATZSL in the no attack scenario, as shown in the four figures.

\begin{table*}[!t]
\caption{Comparative Results ($\textrm{H}_{\textrm{T1}}$:\%) of Several ZSL Methods on \textbf{CUB} under the \textbf{Standard} ZSL with Various Attacks in the \textbf{Semantic} Space.
 The Number in Parentheses Measures the Improvement Obtained by Our ATZSL over the Strongest Competitor in Terms of $\textrm{H}_{\textrm{T1}}$
 } \label{tab:CUB_SZSL_Semanticl}
 \vspace{-0.3cm}
 \begin{center}
  \begin{tabular}{c|c|cccccc}
    \hline \hline
    \textbf{Attack type} &\textbf{Magnitude} & \textbf{DEM}~\cite{zhang2017learning}& \textbf{FGN}~\cite{xian2018feature} & \textbf{MC-ZSL}~\cite{felix2018multi} & \textbf{LDF}~\cite{li2018discriminative} & \textbf{Baseline}& \textbf{ATZSL}(Ours)\\
    \hline
    \multirow{3}*{\textbf{FGSM~\cite{goodfellow2014explaining}}} & $\rho=18$ &  19.9&	22.9&	19.8&	22.1&	26.3&	\textbf{42.2}(15.9)\\ 
    & $\rho=36$ & 13.7	&6.4	&7.8	&9.0	&11.1&	\textbf{31.2 }(17.5)\\
    & $\rho=54$ &0.6&	2.5	&2.7&	5.3	&5.5&	\textbf{24.5}(19.0)\\
    \hline
    \multirow{3}*{\textbf{IFGSM~\cite{kurakin2016adversarial}}} & $\rho=18$ &  16.9&	18.9&	16.1&	16.9&	21.7&	\textbf{40.2} (18.5)\\
    & $\rho=36$ & 4.4&	9.9	&5.2&	2.5	&7.3&	\textbf{25.9} (16.0)\\
    & $\rho=54$ &  1.2&	2.0	&0.8&	2.2&	3.7	&\textbf{18.3}(14.6)\\
    \hline
    \multirow{3}*{\textbf{WRM~\cite{sinha2017certifying}}} & 
    $\rho=18$ &  18.9&	14.5&	15.0&	17.5&	17.7&	\textbf{40.5}(21.6) \\
    & $\rho=36$ & 5.3&	3.5	&10.7&	11.2&	  7.3&  \textbf{38.0}(26.8)\\
    & $\rho=54$ & 2.2&	0.6	&1.4&	2.2	&3.9&	\textbf{27.8}(23.9) \\
    \hline\hline
    \end{tabular}
 \end{center}
\end{table*}

\begin{figure*}[tbp]
\centering
\subfigure[Image attack on AWA2]{\label{image_gzsl_awa}
\begin{minipage}[t]{0.25\linewidth}
\centering
\includegraphics[width=1.75in]{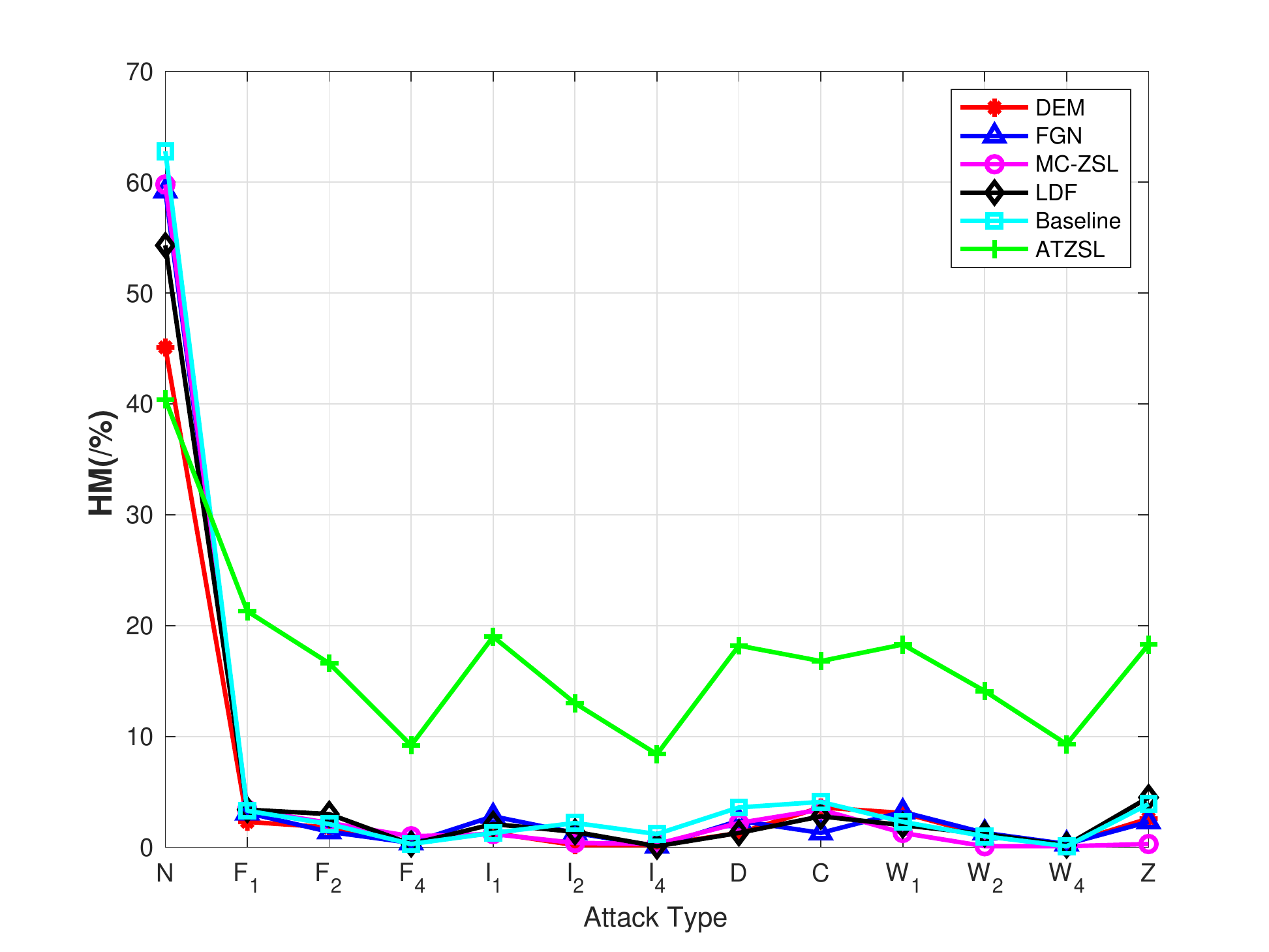}
\end{minipage}%
}%
\subfigure[Attribute attack on AWA2]{\label{att_gzsl_awa}
\begin{minipage}[t]{0.25\linewidth}
\centering
\includegraphics[width=1.75in]{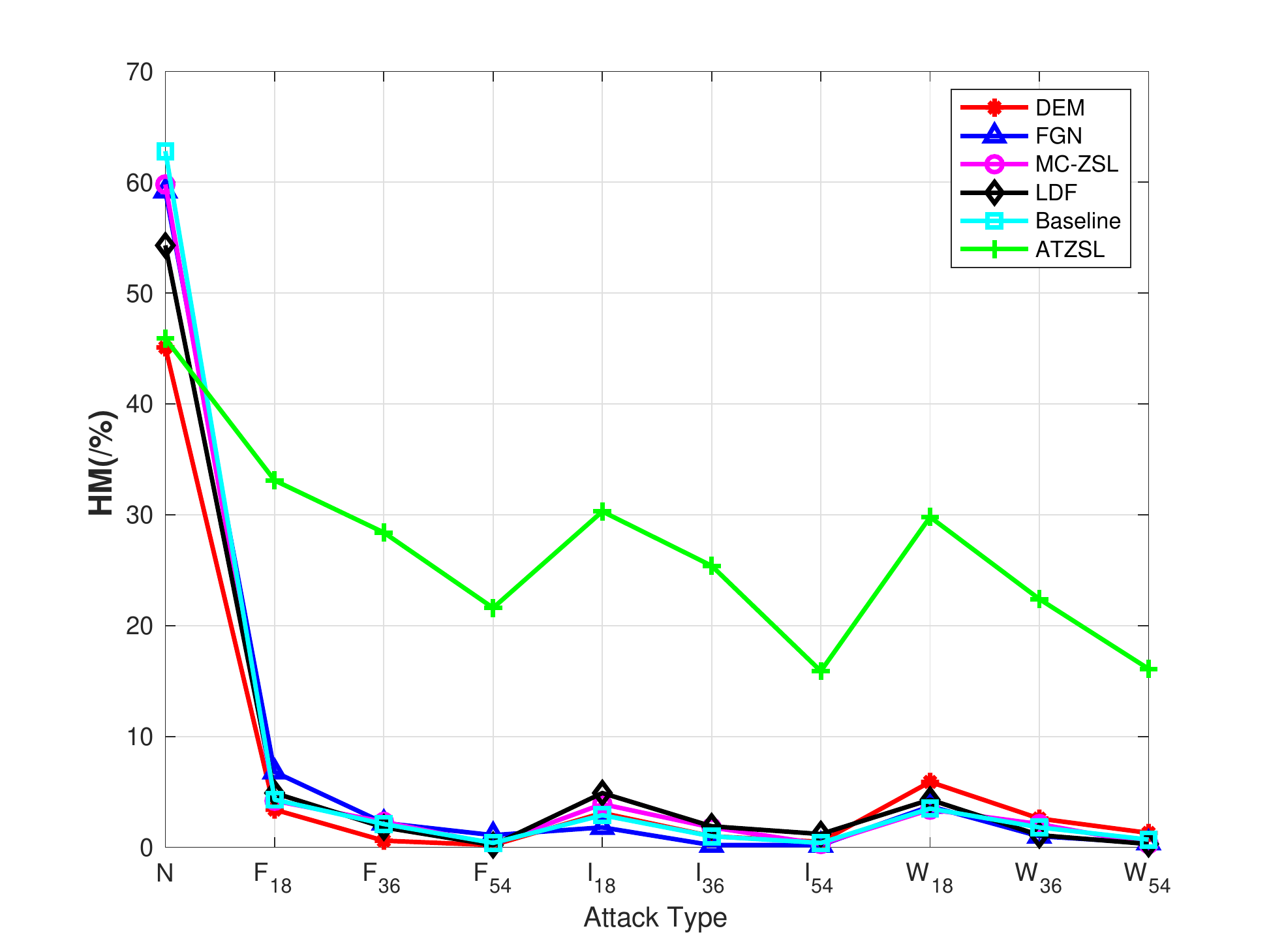}
\end{minipage}%
}%
\subfigure[Image attack on CUB]{\label{image_gzsl_cub}
\begin{minipage}[t]{0.25\linewidth}
\centering
\includegraphics[width=1.75in]{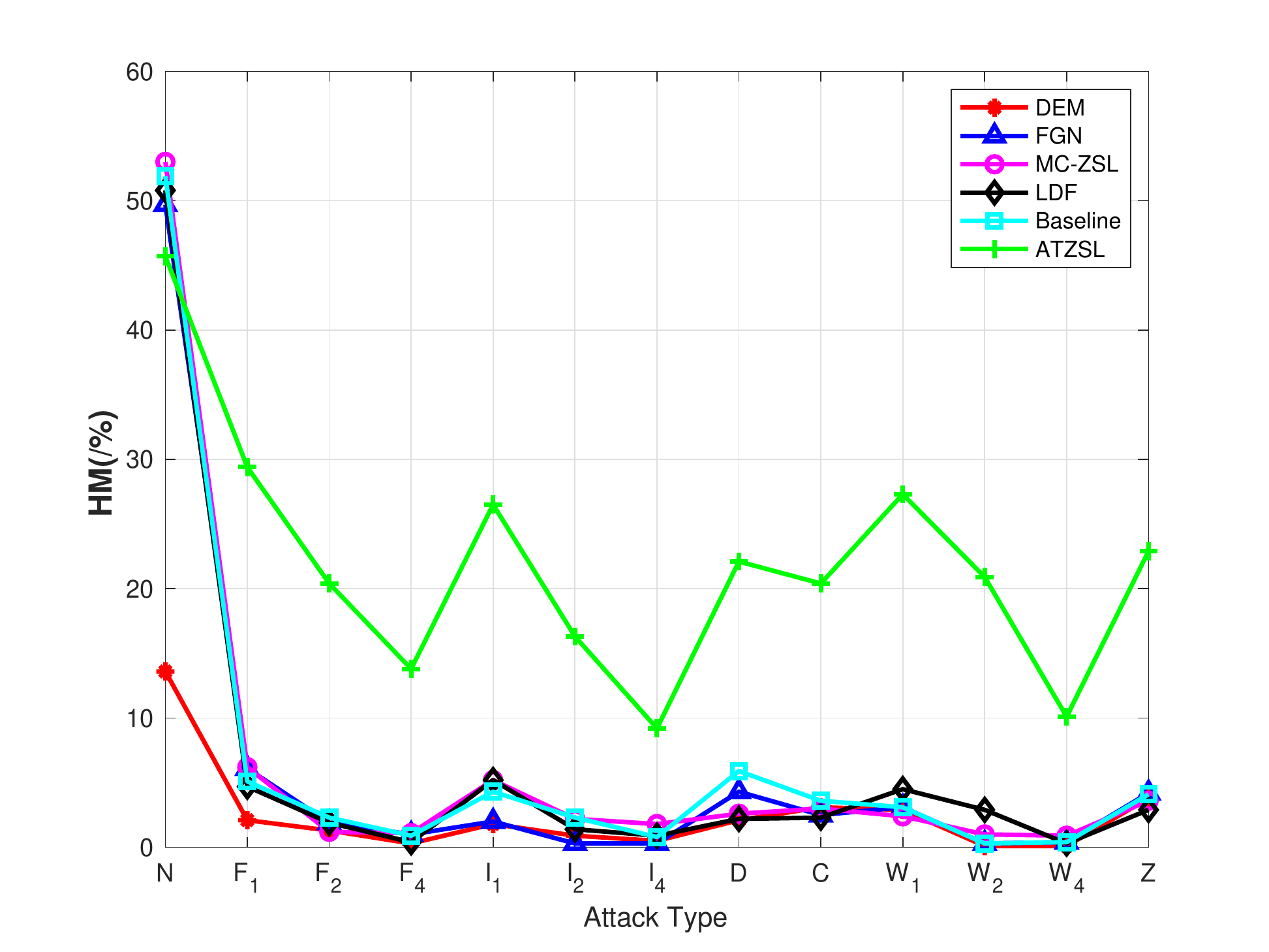}
\end{minipage}
}%
\subfigure[Attribute attack on CUB]{\label{att_gzsl_cub}
\begin{minipage}[t]{0.25\linewidth}
\centering
\includegraphics[width=1.75in]{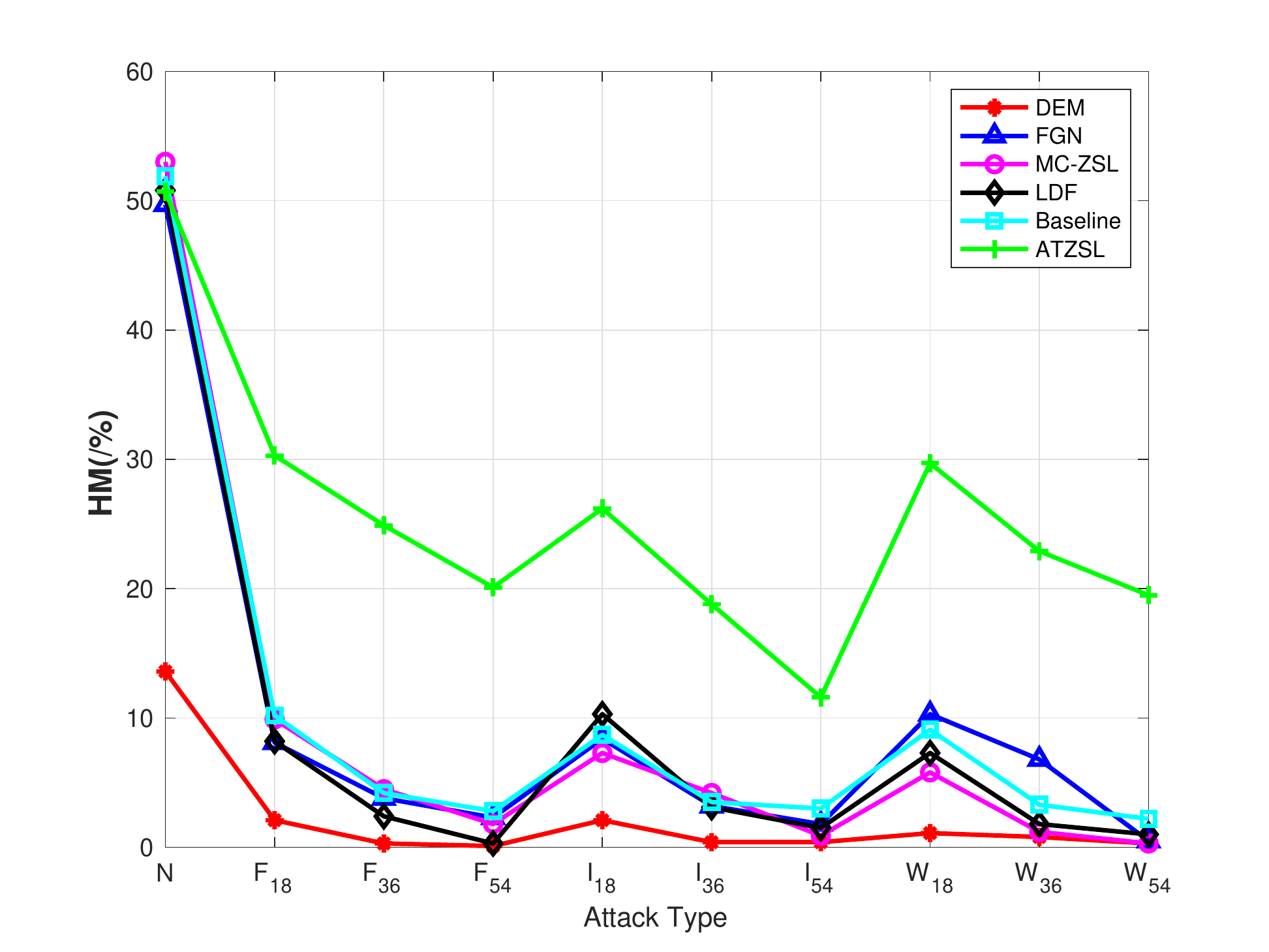}
\end{minipage}
}%
\centering
\caption{Comparison (HM:\%) of several ZSL methods on two datasets under the \textbf{generalized} ZSL setting in various attack scenarios. For horizontal axis, 'N', 'F', 'I', 'D', 'C', 'W', and 'Z' represent No attack, FGSM, IFGSM, DeepFool, CW, WRM, and ZOO attack methods, respectively. The numbers with respect to 'F', 'I', and 'W' represent the attack magnitudes.}
\label{fig:GZSL}
\end{figure*} 

\begin{table*}[!t]
\caption{Comparative Results ($\textrm{H}_{\textrm{HM}}$:\%) of Several ZSL Methods on \textbf{AWA2} under the \textbf{Generalized} ZSL with Various Attacks in the \textbf{Visual} Space. 
'-' Represents that the Attack Magnitude Is Not Required for That Attack Method. 
 The Number in Parentheses Measures the Improvement Obtained by Our ATZSL over the Strongest Competitor in Terms of $\textrm{H}_{\textrm{HM}}$
 } \label{tab:AWA2_GZSL_Visual}
 \vspace{-0.3cm}
 \begin{center}
  \begin{tabular}{c|c|cccccc}
    \hline \hline
    \textbf{Attack type} &\textbf{Magnitude} & \textbf{DEM}~\cite{zhang2017learning}& \textbf{FGN}~\cite{xian2018feature} & \textbf{MC-ZSL}~\cite{felix2018multi} & \textbf{LDF}~\cite{li2018discriminative} & \textbf{Baseline}& \textbf{ATZSL}(Ours)\\
    \hline
    \multirow{3}*{\textbf{FGSM~\cite{goodfellow2014explaining}}} & $\rho=1$ & 4.4&	5.9&	6.4&	6.4&	6.3&	\textbf{27.9}(21.5)\\ 
    & $\rho=2$ & 3.5&	2.7&	4.2&	5.7&	4.1	&  \textbf{23.5}(17.8) \\
    & $\rho=4$ & 0.8&	0.8&	2.0&	0.6&	0.6&	\textbf{15.0}(13.0)\\
    \hline
    \multirow{3}*{\textbf{IFGSM~\cite{kurakin2016adversarial}}} & $\rho=1$ & 2.5&	5.3	&2.4&	4.0&	2.5&	\textbf{25.8}(20.5) \\
    & $\rho=2$ &0.4&	2.5&	0.8&2.7	&4.3&	\textbf{19.7}(15.4)\\
    & $\rho=4$ & 0.4&	0.2	&0.6&	0.2	&2.4&	\textbf{13.9}(11.5)\\
    \hline
    \textbf{DeepFool~\cite{moosavi2016deepfool}} &- & 2.5&	4.6&	4.2&	2.5	&6.8&	\textbf{25.1}(18.3)\\
    \hline
    \textbf{CW~\cite{carlini2017towards}}&-  & 6.7&	2.5	&6.4&	5.3	&7.7&	\textbf{23.7} (16.0) \\
    \hline
    \multirow{3}*{\textbf{WRM~\cite{sinha2017certifying}}} & $\rho=1$ & 5.8	&6.1&	2.5&	3.9&	4.4	&\textbf{25.2} (19.1)\\
    & $\rho=2$ &  2.1&	2.5	&0.2&	2.3	&2.0&	\textbf{20.9}(18.4)\\
    & $\rho=4$ & 0.6&	0.6	&0.2&	0.4&	0.2	&\textbf{15.1}(14.5)\\
    \hline
    \textbf{ZOO~\cite{chen2017zoo}} & -& 4.9&	4.4&	0.6&	8.3&	7.3&	\textbf{25.2}(16.9) \\
    \hline\hline
    \end{tabular}
 \end{center}
\end{table*}

\subsection{Generalized ZSL} \label{sec:gzsl}
Compared with standard ZSL, it is much more technically difficult to perform generalized ZSL since prediction needs to be made over both seen and unseen classes. Even so, we still perform generalized ZSL tasks using several ZSL methods on both AWA2 and CUB datasets, under various attack settings. In particular, we can directly use the ATZSL model trained for standard ZSL tasks in Section~\ref{sec:szsl} for generalized ZSL during the test phase.

\begin{table*}[!t]
\caption{Comparative Results ($\textrm{H}_{\textrm{HM}}$:\%) of Several ZSL Methods on \textbf{AWA2} under the \textbf{Generalized} ZSL with Various Attacks in the \textbf{Semantic} Space.
 The Number in Parentheses Measures the Improvement Obtained by Our ATZSL over the Strongest Competitor in Terms of $\textrm{H}_{\textrm{HM}}$
 } \label{tab:AWA2_GZSL_Semanticl}
 \vspace{-0.3cm}
 \begin{center}
  \begin{tabular}{c|c|cccccc}
    \hline \hline
    \textbf{Attack type} &\textbf{Magnitude} & \textbf{DEM}~\cite{zhang2017learning}& \textbf{FGN}~\cite{xian2018feature} & \textbf{MC-ZSL}~\cite{felix2018multi} & \textbf{LDF}~\cite{li2018discriminative} & \textbf{Baseline}& \textbf{ATZSL}(Ours)\\
    \hline
    \multirow{3}*{\textbf{FGSM~\cite{goodfellow2014explaining}}} & $\rho=9$ & 6.3 &	12.2 &	7.8 &	9.0 &	8.0 &	\textbf{38.5}(26.3)\\ 
    & $\rho=18$ & 1.2 &	4.2 &	4.4 &	3.5	 &4.1 &	\textbf{35.1}(30.7) \\
    & $\rho=27$ & 0.4 &	2.2 &	0.6	 &0.4	 &0.8 &	\textbf{29.4}(27.2)\\
    \hline
    \multirow{3}*{\textbf{IFGSM~\cite{kurakin2016adversarial}}} & $\rho=9$ & 5.8 &	3.5 &	7.3 &	9.0 &	5.5 &	\textbf{36.5}(27.5)\\
    & $\rho=18$ &  2.0	 &0.4 &	3.5	 &3.7 &	2.0	 &\textbf{32.7}(29.0) \\
    & $\rho=27$ & 1.0 &	0.4	 &0.6 &	2.3 &	0.8	 &\textbf{23.6}(21.3)\\
    \hline
    \multirow{3}*{\textbf{WRM~\cite{sinha2017certifying}}} & 
    $\rho=9$ &  10.4 &	7.0	 &6.4 &	8.0 &	6.6	 &\textbf{36.1}(25.7) \\
    & $\rho=18$ & 4.9 &	2.0	 &4.1 &	2.2	 &3.5	 &\textbf{30.1}(25.2)\\
    & $\rho=27$ & 2.5 &	0.8 &	0.6 &	0.6 &	1.4	 &\textbf{23.8}(21.3) \\
    \hline\hline
    \end{tabular}
 \end{center}
\end{table*}

\begin{table*}[!t]
 \caption{Comparative Results ($\textrm{H}_{\textrm{HM}}$:\%) of Several ZSL Methods on \textbf{CUB} under the \textbf{Generalized} ZSL with Various Attacks in the \textbf{Visual} Space.  
 '-' Represents that the Attack Magnitude Is Not Required for That Attack Method. 
 The Number in Parentheses Measures the Improvement Obtained by Our ATZSL over the Strongest Competitor in Terms of $\textrm{H}_{\textrm{HM}}$
 } \label{tab:CUB_GZSL_Visual}
 \vspace{-0.3cm}
 \begin{center}
  \begin{tabular}{c|c|cccccc}
    \hline \hline
    \textbf{Attack type} &\textbf{Magnitude} & \textbf{DEM}~\cite{zhang2017learning}& \textbf{FGN}~\cite{xian2018feature} & \textbf{MC-ZSL}~\cite{felix2018multi} & \textbf{LDF}~\cite{li2018discriminative} & \textbf{Baseline}& \textbf{ATZSL}(Ours)\\
    \hline
    \multirow{3}*{\textbf{FGSM~\cite{goodfellow2014explaining}}} & $\rho=1$ &3.6&	10.9&	11.1&	8.6&	9.3	&\textbf{35.8}(24.7)\\ 
    & $\rho=2$ &  2.4&	3.7	&2.3&	3.7	&4.4&	\textbf{28.2}(23.8) \\
    & $\rho=4$ & 0.6&	2.0	&2.2&	0.8	&1.8	&\textbf{21.2}(19.0)\\
    \hline
    \multirow{3}*{\textbf{IFGSM~\cite{kurakin2016adversarial}}} & $\rho=1$ & 3.2&	3.8&	9.5&	9.4	&7.9&	\textbf{33.5}(24.0) \\
    & $\rho=2$ & 1.7&	0.6&   4.2& 2.7&	4.4&	\textbf{24.0}(19.6) \\
    & $\rho=4$ & 1.0&	0.6&   3.5&	1.8&    1.6&    \textbf{15.3}(11.8) \\
    \hline
    \textbf{DeepFool~\cite{moosavi2016deepfool}} &- & 3.6&	7.9&5.0&4.2&10.6&	\textbf{29.8 }(19.2)\\
    \hline
    \textbf{CW~\cite{carlini2017towards}}&-  & 5.0&	4.8	&5.7&	4.4&	6.7&	\textbf{28.2}(21.5)\\
    \hline
    \multirow{3}*{\textbf{WRM~\cite{sinha2017certifying}}} & $\rho=1$ & 4.8	&5.8&	4.6&	8.3&	5.9	&\textbf{34.2} (25.9)\\
    & $\rho=2$ &  0.2&	0.6&2.0	&5.5&	0.6	&\textbf{28.7 }(23.2)\\
    & $\rho=4$ &  0.2&	0.8	&1.8&	0.6&	0.8	&\textbf{16.5}(14.7)\\
    \hline
    \textbf{ZOO~\cite{chen2017zoo}} & -& 5.9&	7.7	&6.7&	5.5	&7.6&	\textbf{30.5 }(22.8)\\
    \hline\hline
    \end{tabular}
 \end{center}
\end{table*}

\begin{table*}[!t]
\caption{Comparative Results ($\textrm{H}_{\textrm{HM}}$:\%) of Several ZSL Methods on \textbf{CUB} under the \textbf{Generalized} ZSL with Various Attacks in the \textbf{Semantic} Space.
The Number in Parentheses Measures the Improvement Obtained by Our ATZSL over the Strongest Competitor in Terms of $\textrm{H}_{\textrm{HM}}$
 } \label{tab:CUB_GZSL_Semanticl}
 \vspace{-0.3cm}
 \begin{center}
  \begin{tabular}{c|c|cccccc}
    \hline \hline
    \textbf{Attack type} &\textbf{Magnitude} & \textbf{DEM}~\cite{zhang2017learning}& \textbf{FGN}~\cite{xian2018feature} & \textbf{MC-ZSL}~\cite{felix2018multi} & \textbf{LDF}~\cite{li2018discriminative} & \textbf{Baseline}& \textbf{ATZSL}(Ours)\\
    \hline
    \multirow{3}*{\textbf{FGSM~\cite{goodfellow2014explaining}}} & $\rho=18$ & 3.6& 	13.9& 	16.7& 	14.1& 	17.0& 	\textbf{37.9}(20.9)\\ 
    & $\rho=36$ & 0.6& 	7.1	& 8.3& 	4.6& 	7.8	&\textbf{ 33.4}(25.1) \\
    & $\rho=54$ & 0.2& 	4.4& 	3.5	& 0.6 & 	5.3	& \textbf{28.8}(23.5)\\
    \hline
    \multirow{3}*{\textbf{IFGSM~\cite{kurakin2016adversarial}}} & $\rho=18$ &  3.6& 	14.4& 	12.8& 	17.1& 	14.9& 	\textbf{34.5}(17.4)\\
    & $\rho=36$ & 0.8& 	6.0	& 7.8& 	5.8& 	6.6& 	\textbf{27.4} (19.6)\\
    & $\rho=54$ & 0.8& 	3.5	& 1.8& 	2.9	& 5.7& 	\textbf{18.9}(13.2)\\
    \hline
    \multirow{3}*{\textbf{WRM~\cite{sinha2017certifying}}} & $\rho=18$ &  2.0& 	17.1& 	10.5& 	12.8& 	15.5& 	\textbf{37.5} (20.4)\\
    & $\rho=36$ & 1.5& 	8.8	& 2.3& 	3.5& 	6.2& \textbf{31.5}(22.7)\\
    & $\rho=54$ & 0.6& 	1.0	& 0.6& 	2.0& 	4.2& 	\textbf{28.2 }(24.0)\\
    \hline\hline
    \end{tabular}
 \end{center}
\end{table*}

\textbf{Generalized ZSL Results on AWA2.} 
Under the generalized ZSL setting, we first evaluate the performances of ATZSL in Eq.~(\ref{ATZSL}) and Eq.~(\ref{atta}) on AWA2 dataset, corresponding to the image and attribute attacks respectively. It is worth noting that the two models are trained as those in standard ZSL on AWA2. During the test phase, we need to search in both the seen and the unseen class spaces, where the attack implementation is the same as that in Section~\ref{sec:szsl}. 
After attacking images in AWA2 with six types of attack methods and three types of magnitudes, we show the compared results in terms of harmonic mean ($\textrm{HM}$) with five ZSL methods in Fig.~\ref{image_gzsl_awa}. Concretely, `N' in horizontal axis still denotes the performance without any attacks (i.e., $\textrm{HM}_{\textrm{cle}}$), and others represent $\textrm{HM}_{\textrm{adv}}$. 
Obviously, unlike those ZSL baselines that are very sensitive to image attacks, our ATZSL achieves the slightest degradation in various image attack scenarios. 
Meanwhile, Table~\ref{tab:AWA2_GZSL_Visual} reports the harmonic mean ($\textrm{H}_{\textrm{HM}}$) of $\textrm{HM}_{\textrm{cle}}$ and $\textrm{HM}_{\textrm{adv}}$ under each attack method, i.e., the trade-off of each method on clean and adversarial images.
It can be seen that our ATZSL can also perform best on the generalized ZSL task even with various image attacks. The improvements over the strongest competitor range from 11.5\% to 21.5\%. 
Moreover, by attacking attributes in AWA2 with three kinds of attack methods and three types of magnitudes, Fig.~\ref{att_gzsl_awa} plots the compared results in terms of harmonic mean ($\textrm{HM}$), including $\textrm{HM}_{\textrm{cle}}$ and $\textrm{HM}_{\textrm{adv}}$ for each attack method. Their harmonic means ($\textrm{H}_{\textrm{HM}}$) are computed in Table~\ref{tab:AWA2_GZSL_Semanticl}. 
It is as expected that our ATZSL has more promising trade-off performance on clean and adversarial attributes (generally more than 21.3\%), and meanwhile, only loses a negligible performance on clean attributes as shown in `N' of the horizontal axis of Fig.~\ref{att_gzsl_awa}.

\begin{figure*}[tbp]
\centering
\subfigcapskip=-15pt
\subfigure[$\rho=1$]{\label{cub1}
\includegraphics[width=5.0in]{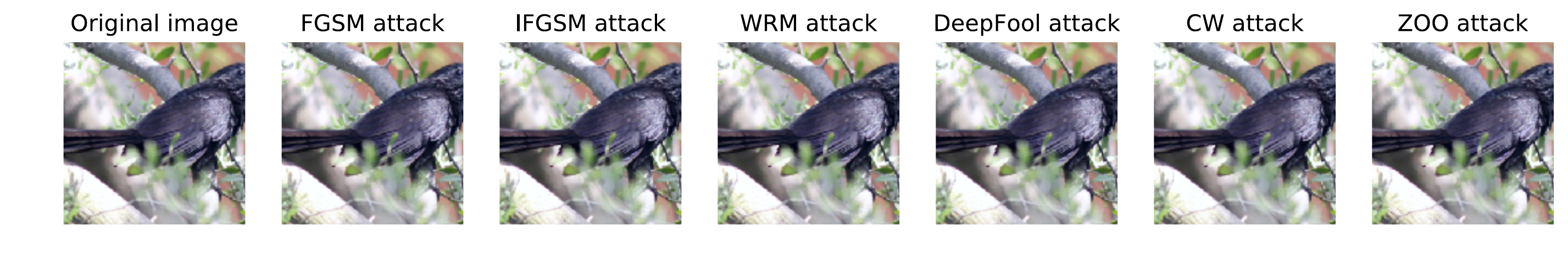}
}%

\subfigcapskip=-15pt 
\subfigure[$\rho=2$]{\label{cub2}
\includegraphics[width=5.0in]{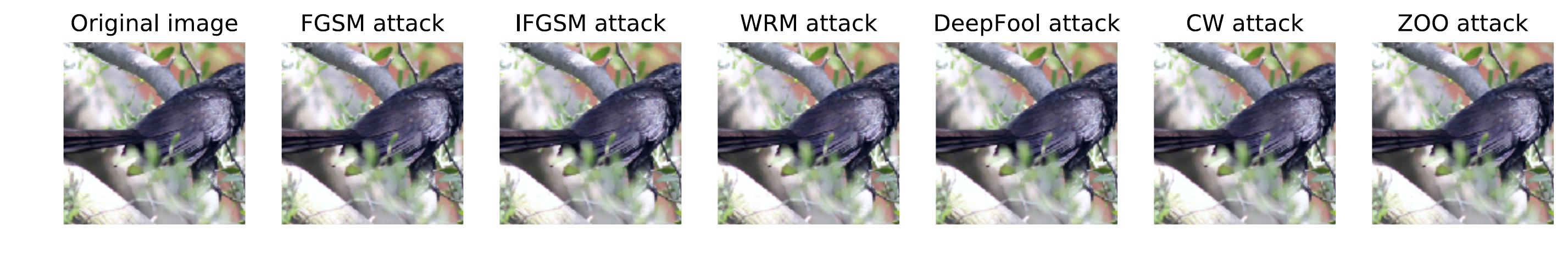}
}%

\subfigcapskip=-15pt 
\subfigure[$\rho=4$]{\label{cub3}
\includegraphics[width=5.0in]{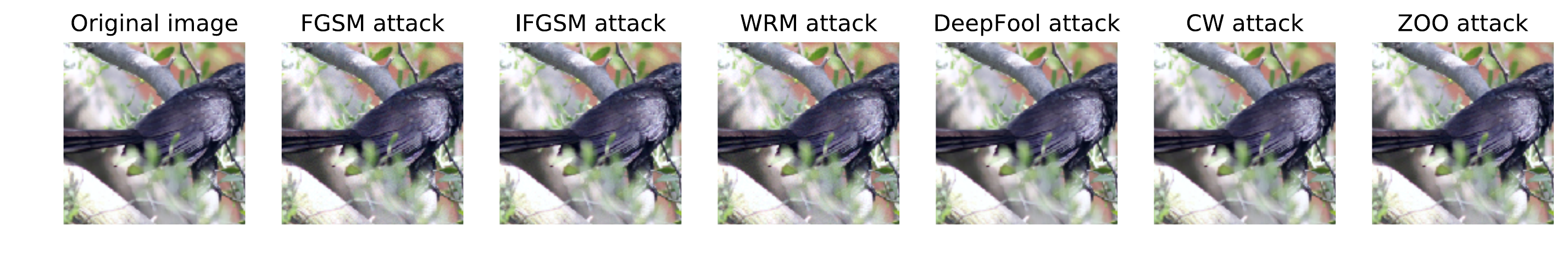}
}%

\subfigcapskip=-15pt 
\subfigure[$\rho=8$]{\label{cub4}
\includegraphics[width=5.0in]{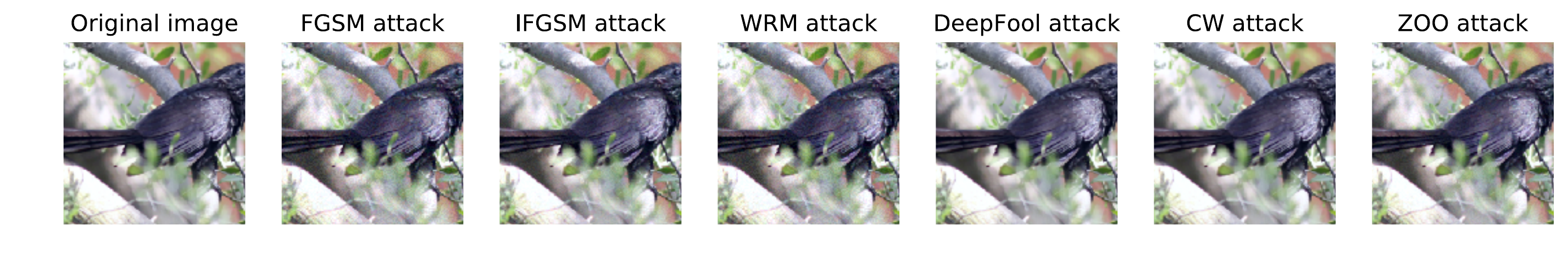}
}%

\subfigcapskip=-15pt 
\subfigure[$\rho=16$]{\label{cub5}
\includegraphics[width=5.0in]{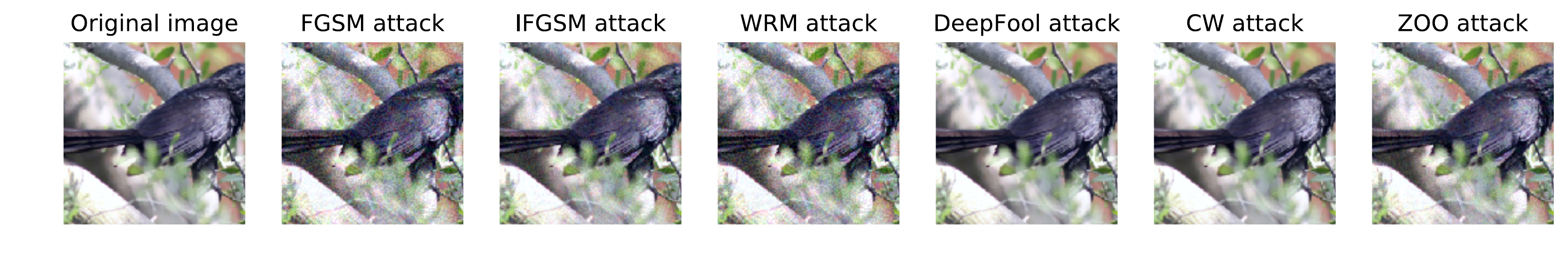}
}%
\centering
\caption{The test unseen images on CUB dataset attacked by various attack methods with five types of attack magnitudes ($\rho$).}
\label{fig:cub-image}
\end{figure*} 

\begin{figure*}[tbp]
\centering
\subfigure[$\rho=18$]{\label{cubatt1}
\includegraphics[width=2.1in]{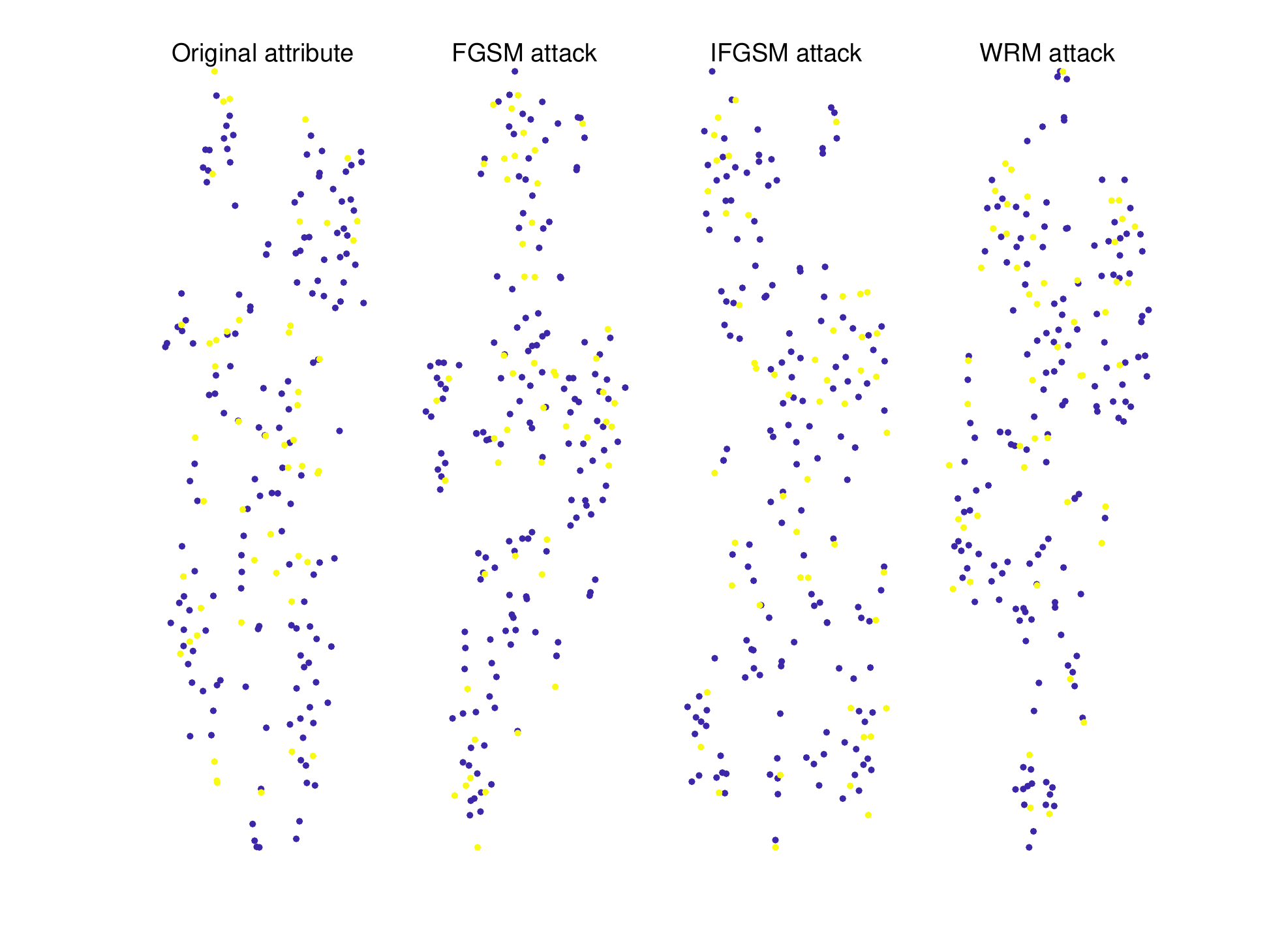}
}%
\subfigure[$\rho=36$]{\label{cubatt2}
\includegraphics[width=2.1in]{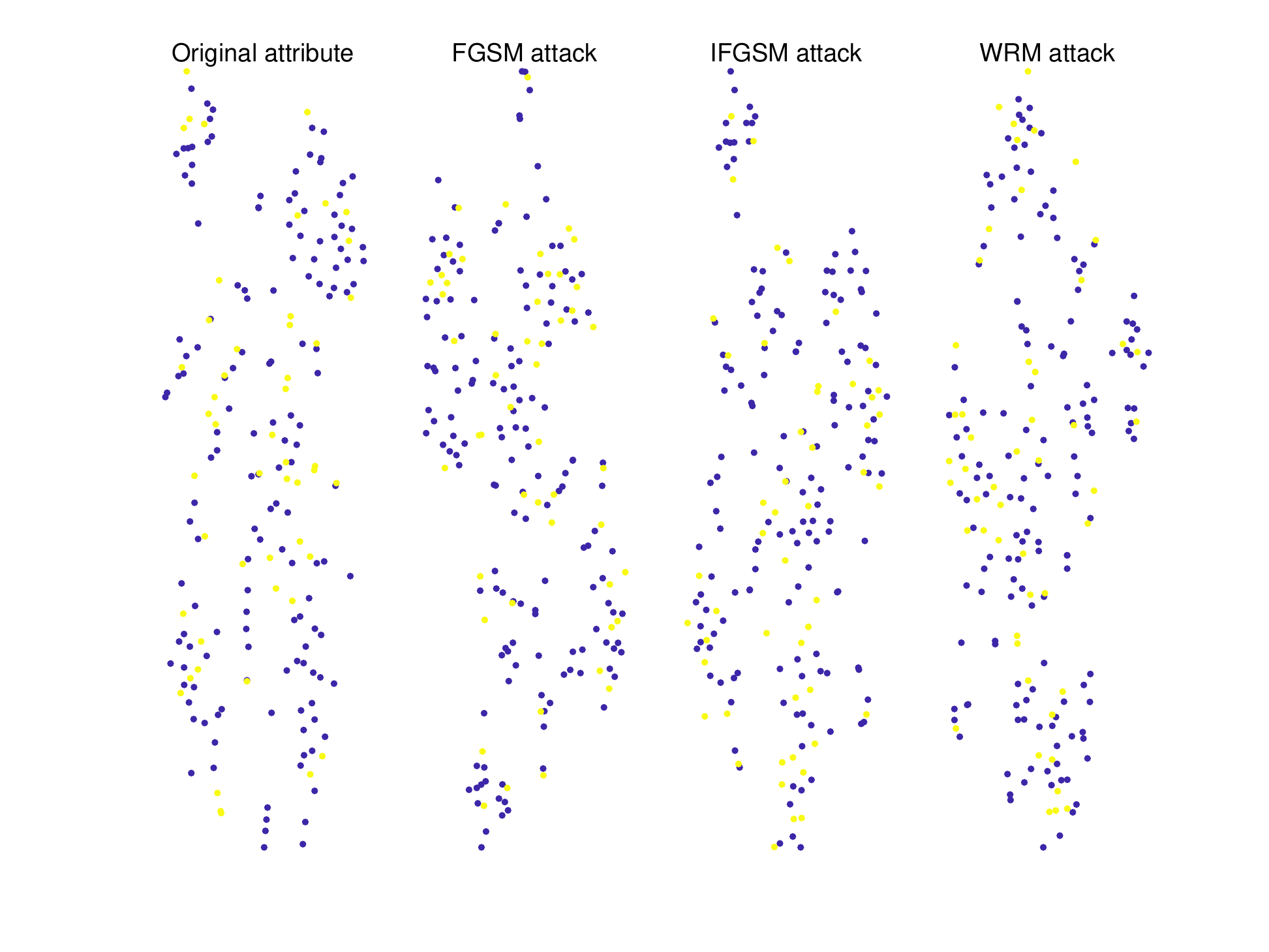}
}%
\subfigure[$\rho=54$]{\label{cubatt3}
\includegraphics[width=2.1in]{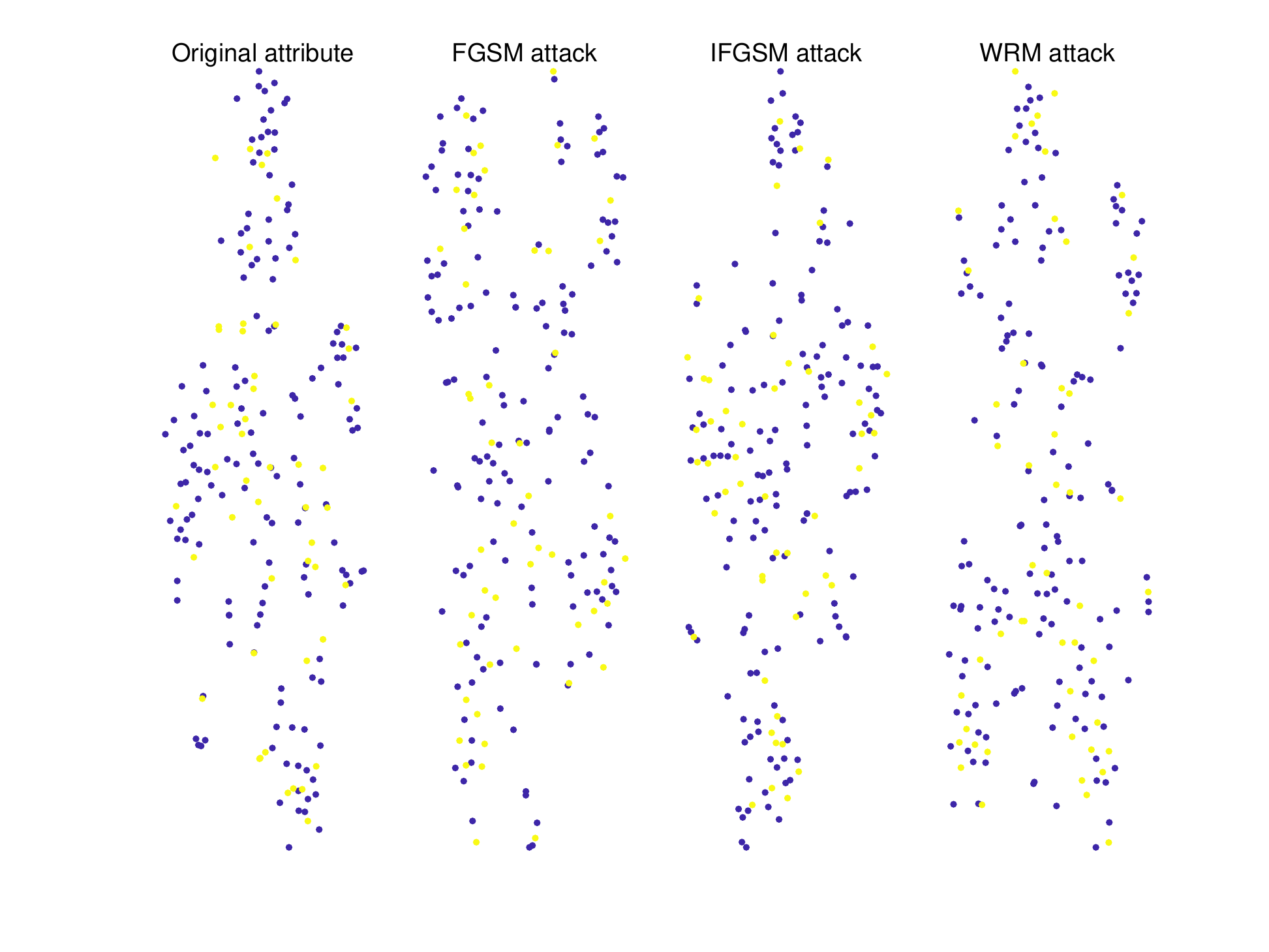}
}%
\centering
\caption{The t-SNE of all class attributes on CUB dataset attacked by various attack methods during the test phase with three types of attack magnitudes ($\rho$). For each figure, the dark color represents the seen class attributes without any attacks, and the light color represents the unseen class attributes with no attacks, and three types of attacks respectively.}
\label{fig:cub-att}
\end{figure*} 

\textbf{Generalized ZSL Results on CUB.} 
Likewise, we further evaluate the generalized ZSL performances of ATZSL in Eq.~(\ref{ATZSL}) and Eq.~(\ref{atta}) on CUB dataset.
By attacking images with various methods, we plot the compared results in terms of harmonic mean ($\textrm{HM}$) in Fig.~\ref{image_gzsl_cub}, including $\textrm{HM}_{\textrm{cle}}$ and $\textrm{HM}_{\textrm{adv}}$ for each attack method. Their harmonic means ($\textrm{H}_{\textrm{HM}}$) are computed in Table~\ref{tab:CUB_GZSL_Visual}.
Then, by attacking attributes with several methods, we plot the compared results in terms of harmonic mean ($\textrm{HM}$) in Fig.~\ref{att_gzsl_cub}, including $\textrm{HM}_{\textrm{cle}}$ and $\textrm{HM}_{\textrm{adv}}$ for each attack method. Their harmonic means ($\textrm{H}_{\textrm{HM}}$) are presented in Table~\ref{tab:CUB_GZSL_Semanticl}.
By contrast, we can make consistent conclusions with generalized ZSL on AWA2.
That is, the robustness and transferability abilities can indeed be jointly optimized better by our ATZSL.
Specifically, as reported in Table~\ref{tab:CUB_GZSL_Visual} and Table~\ref{tab:CUB_GZSL_Semanticl}, even though the unseen images and attributes are attacked, and may be consequently similar with those seen class images and attributes, respectively. ATZSL still can improve the overall performance (i.e., harmonic mean $\textrm{H}_{\textrm{HM}}$) over the strongest alternative by an obvious margin (11.8\% $\sim$ 25.9\% for image attacks, and 13.2\% $\sim$ 25.1\% for attribute attacks, respectively).

Besides the similar conclusion to standard ZSL, we have an additional observation for generalized ZSL. As shown in Fig.~\ref{fig:ZSL} and Fig.~\ref{fig:GZSL}, the performances in terms of $\textrm{HM}$ in generalized ZSL are much weaker than $\textrm{T1}$ in standard ZSL. This is not surprising since the seen classes are included in the search space which act as distractors for the samples that come from unseen classes.

To straightforwardly illustrate the attack effect on images, we show one image in CUB dataset attacked by six kinds of methods during the test phase in Fig.~\ref{fig:cub-image}
corresponding to five attack magnitudes. It can be observed that the magnitude required attack methods (i.e., FGSM~\cite{goodfellow2014explaining}, IFGSM~\cite{kurakin2016adversarial}, and WRM~\cite{sinha2017certifying}) are generally perceptible when $\rho>4$. In addition, by comparing original image with various attacked images, we can find that the gradient based attack methods, including FGSM~\cite{goodfellow2014explaining}, IFGSM~\cite{kurakin2016adversarial}, and WRM~\cite{sinha2017certifying}, are very strong, thus being challenging to defense.
Furthermore, we also illustrate the attribute distribution before and after being attacked using t-SNE~\cite{maaten2008visualizing}. 
As shown in Fig.~\ref{fig:cub-att}
for CUB dataset with three types of attack magnitudes, a small adversarial perturbation is able to change the relationship between the seen and the unseen class attributes. 
This will definitely deteriorate the zero-shot recognition performance by transferring wrong information, if it is not been defensed. 
More visualized results are provided in the supplementary material.
The same conclusion can be summarized from AWA2 dataset. Therefore, the research on robustness of zero-shot learning is very necessary and valuable.


\begin{figure*}[tbp]
\centering
\subfigure[Image attack on AWA2]{\label{image_szsl_awa_aba}
\begin{minipage}[t]{0.25\linewidth}
\centering
\includegraphics[width=1.75in]{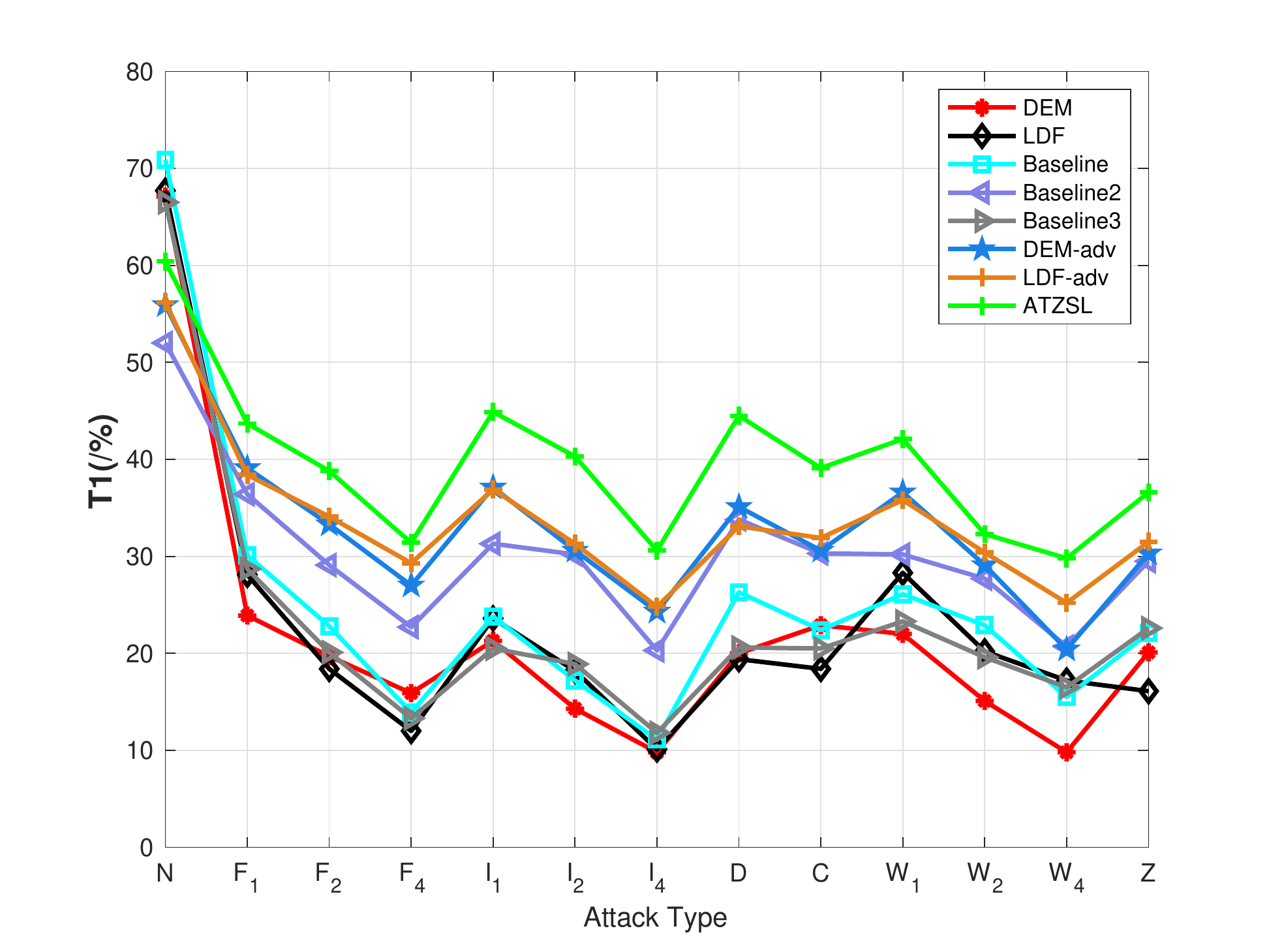}
\end{minipage}%
}%
\subfigure[Attribute attack on AWA2]{\label{att_szsl_awa_aba}
\begin{minipage}[t]{0.25\linewidth}
\centering
\includegraphics[width=1.75in]{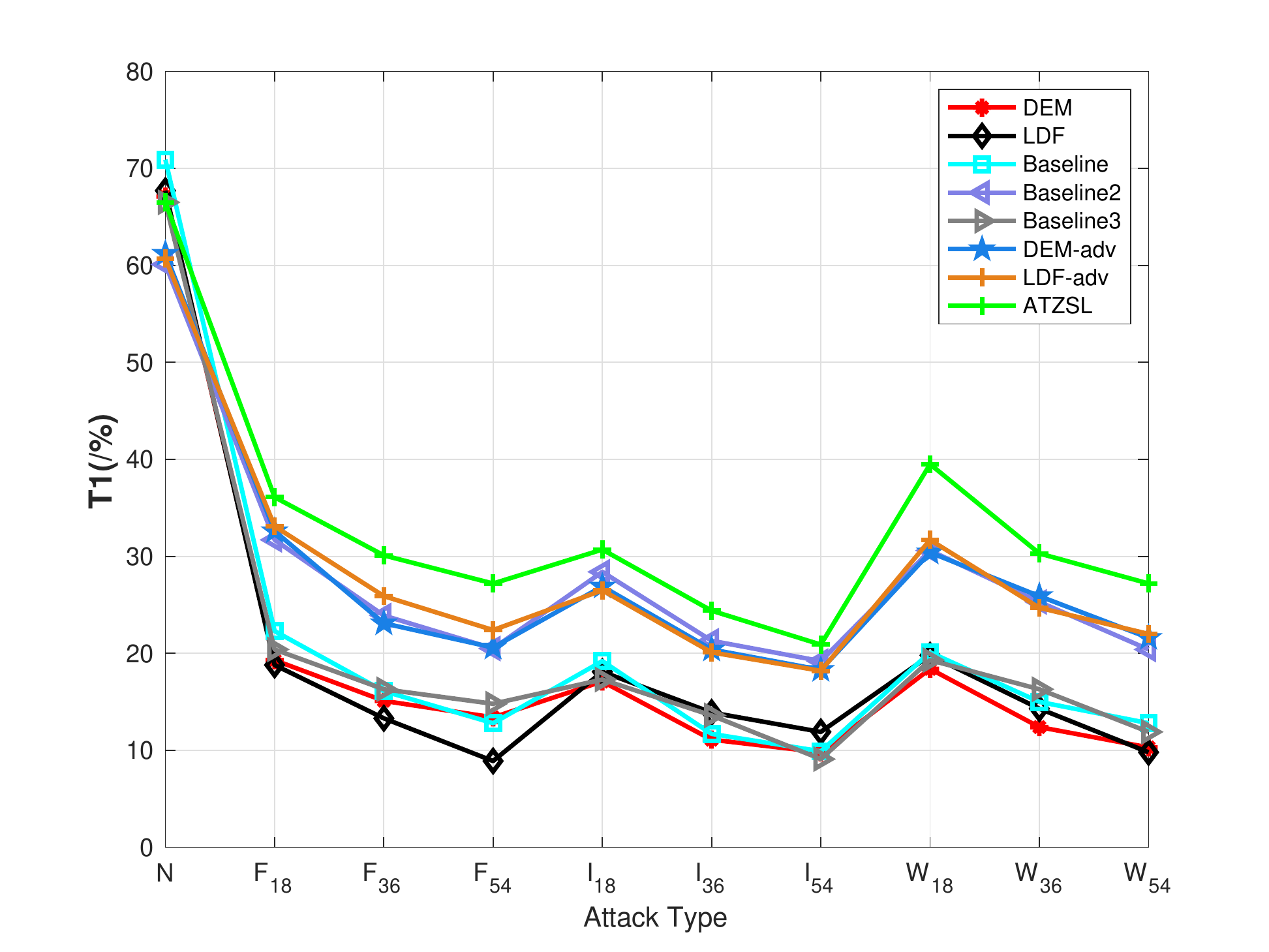}
\end{minipage}%
}%
\subfigure[Image attack on CUB]{\label{image_szsl_cub_aba}
\begin{minipage}[t]{0.25\linewidth}
\centering
\includegraphics[width=1.75in]{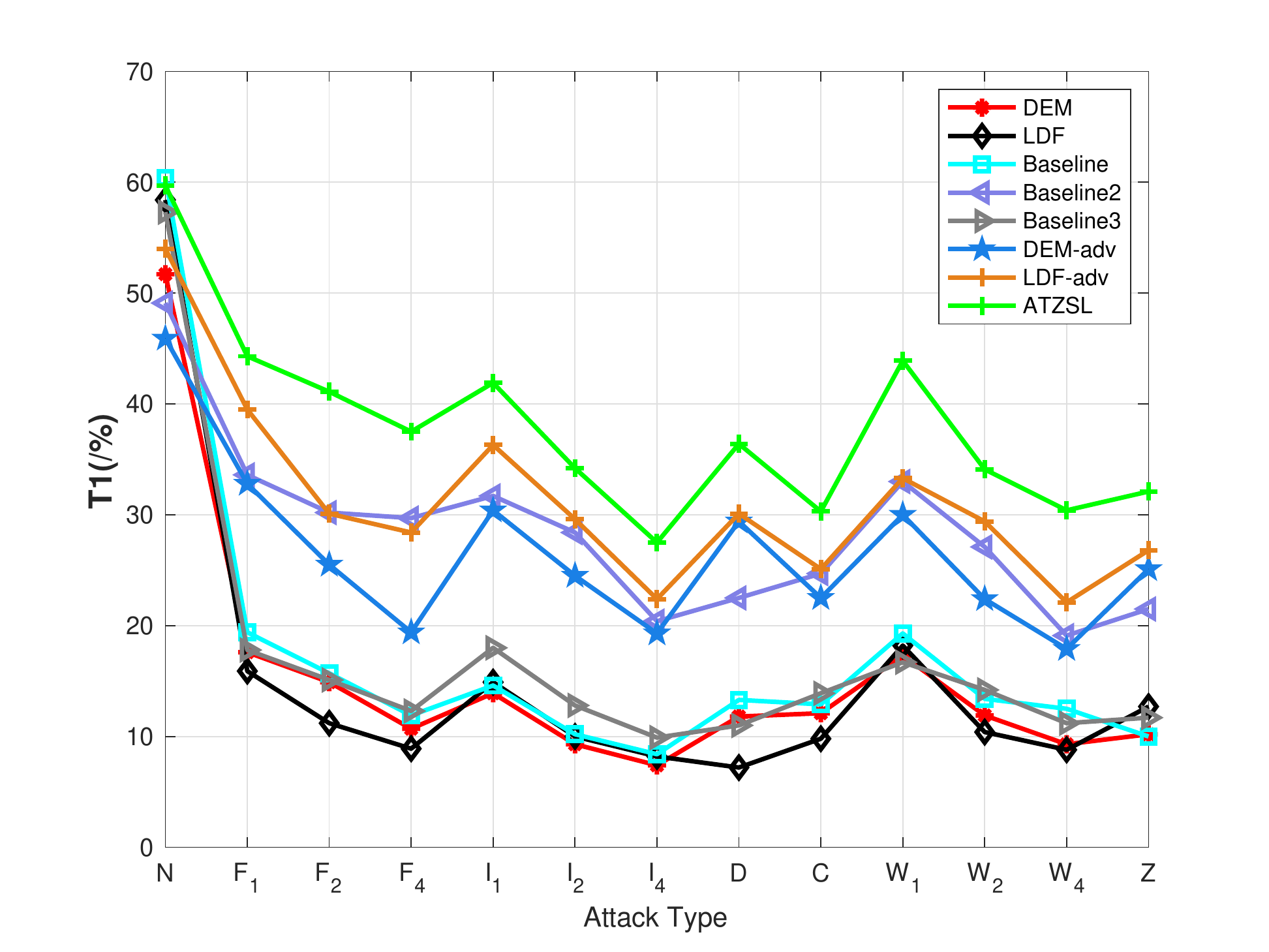}
\end{minipage}
}%
\subfigure[Attribute attack on CUB]{\label{att_szsl_cub_aba}
\begin{minipage}[t]{0.25\linewidth}
\centering
\includegraphics[width=1.75in]{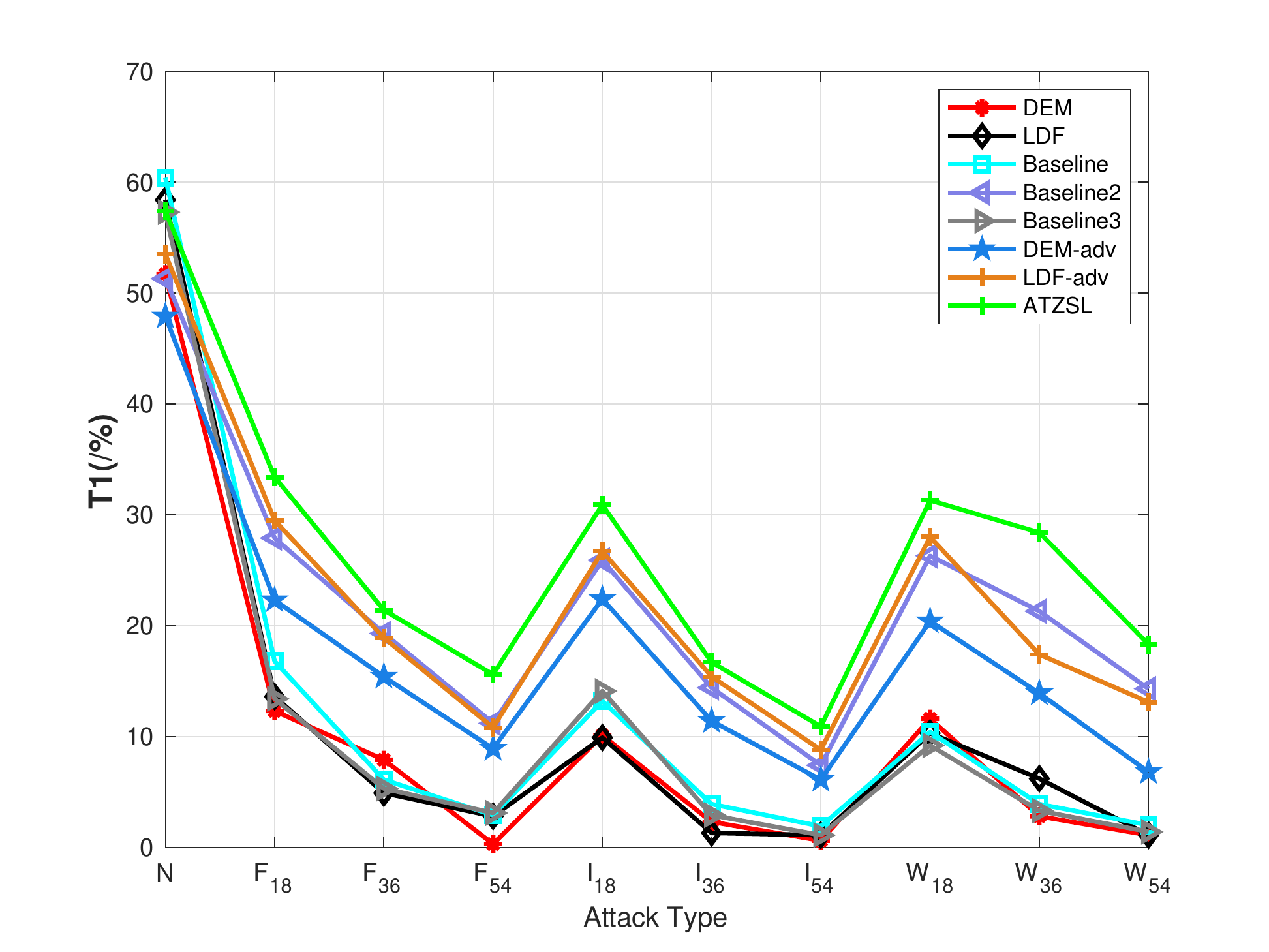}
\end{minipage}
}%
\centering
\caption{Comparison (T1:\%) of several ZSL methods with their adapted versions on two datasets under the \textbf{standard} ZSL setting in various attack scenarios. For horizontal axis, 'N', 'F', 'I', 'D', 'C', 'W', and 'Z' represent No attack, FGSM, IFGSM, DeepFool, CW, WRM, and ZOO attack methods, respectively. The numbers with respect to 'F', 'I', and 'W' represent the attack magnitudes.}
\label{fig:ZSL_aba}
\end{figure*} 

\begin{figure*}[tbp]
\centering
\subfigure[Image attack on AWA2]{\label{image_gzsl_awa_aba}
\begin{minipage}[t]{0.25\linewidth}
\centering
\includegraphics[width=1.75in]{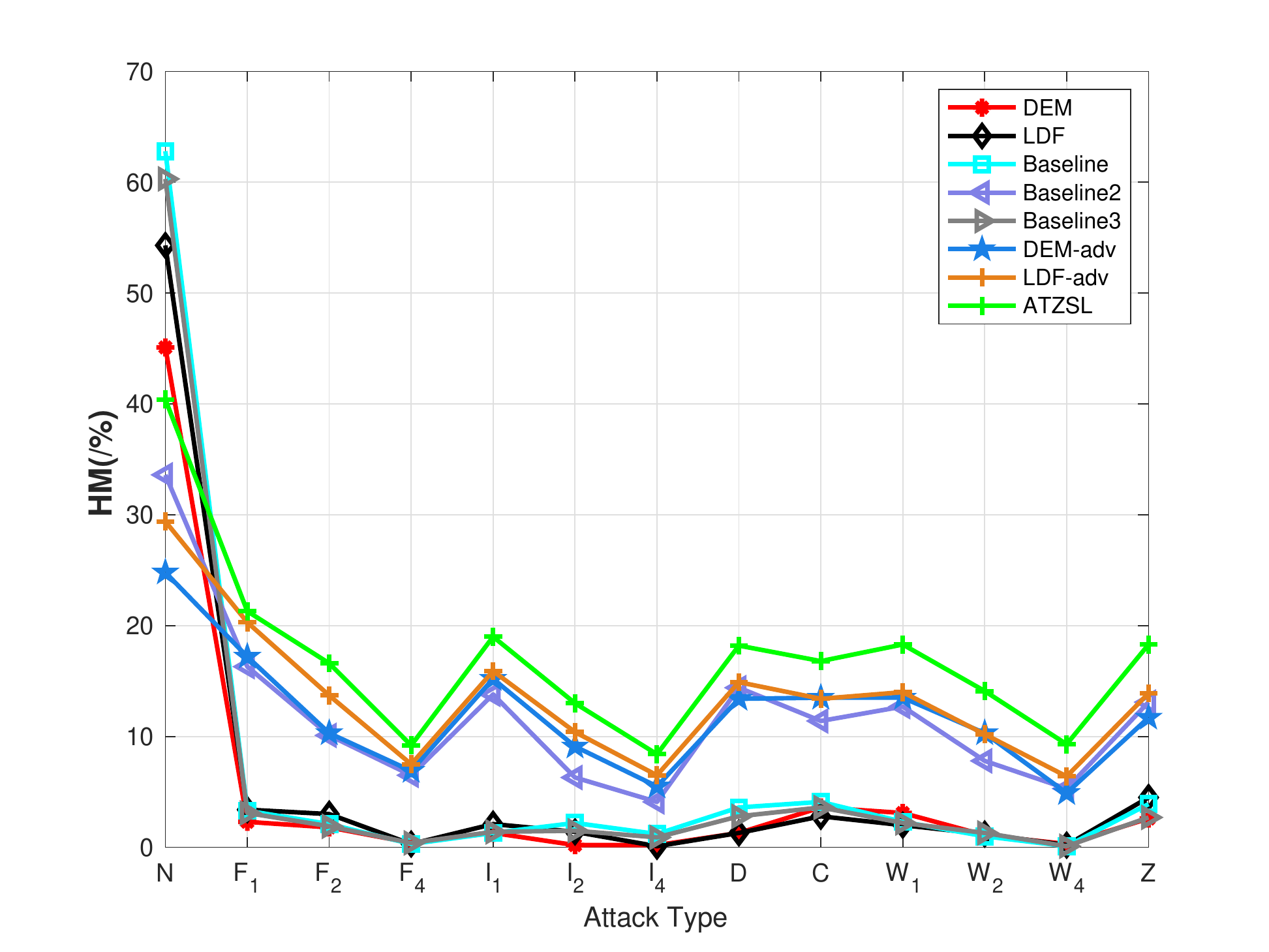}
\end{minipage}%
}%
\subfigure[Attribute attack on AWA2]{\label{att_gzsl_awa_aba}
\begin{minipage}[t]{0.25\linewidth}
\centering
\includegraphics[width=1.75in]{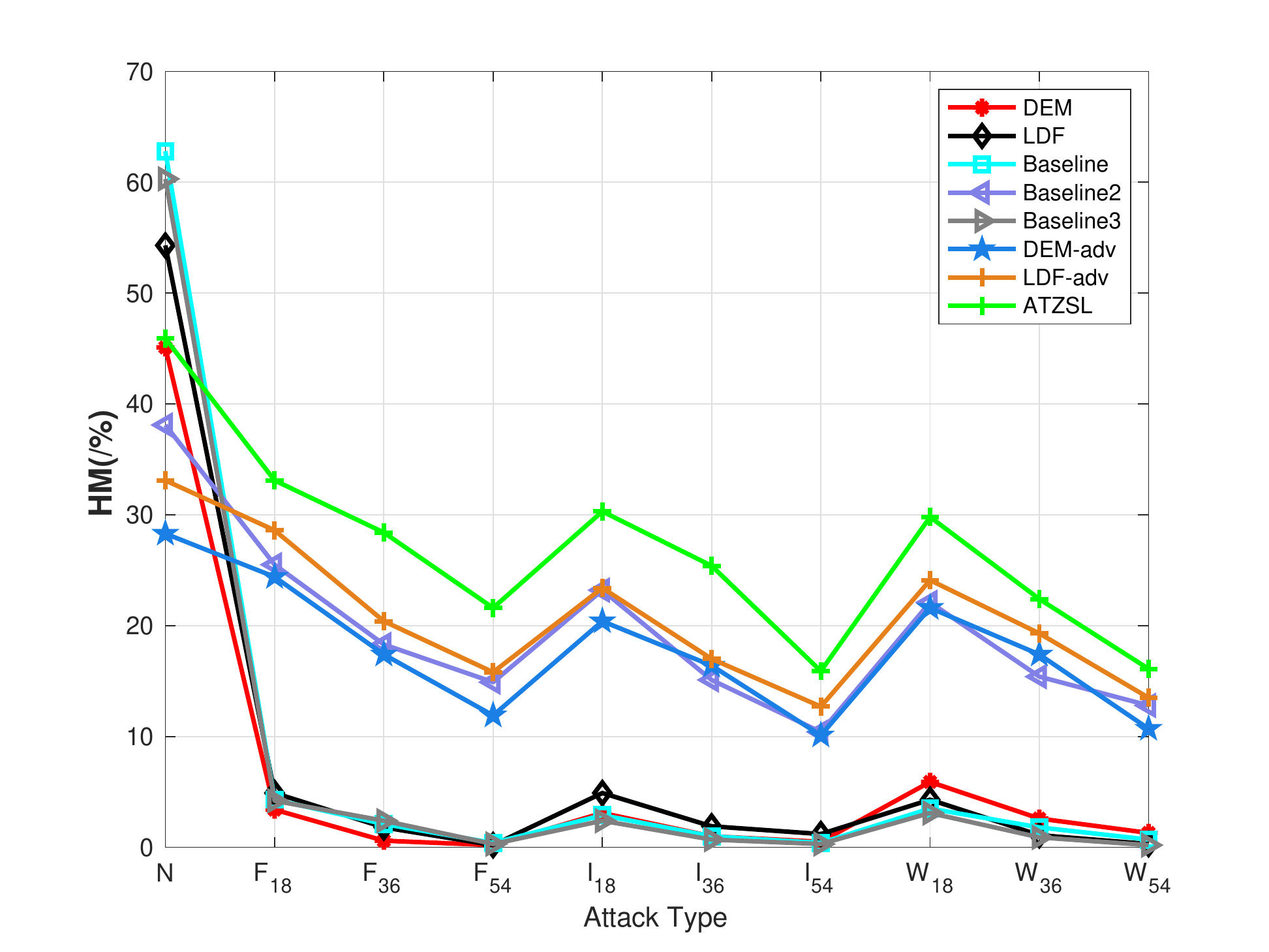}
\end{minipage}%
}%
\subfigure[Image attack on CUB]{\label{image_gzsl_cub_aba}
\begin{minipage}[t]{0.25\linewidth}
\centering
\includegraphics[width=1.75in]{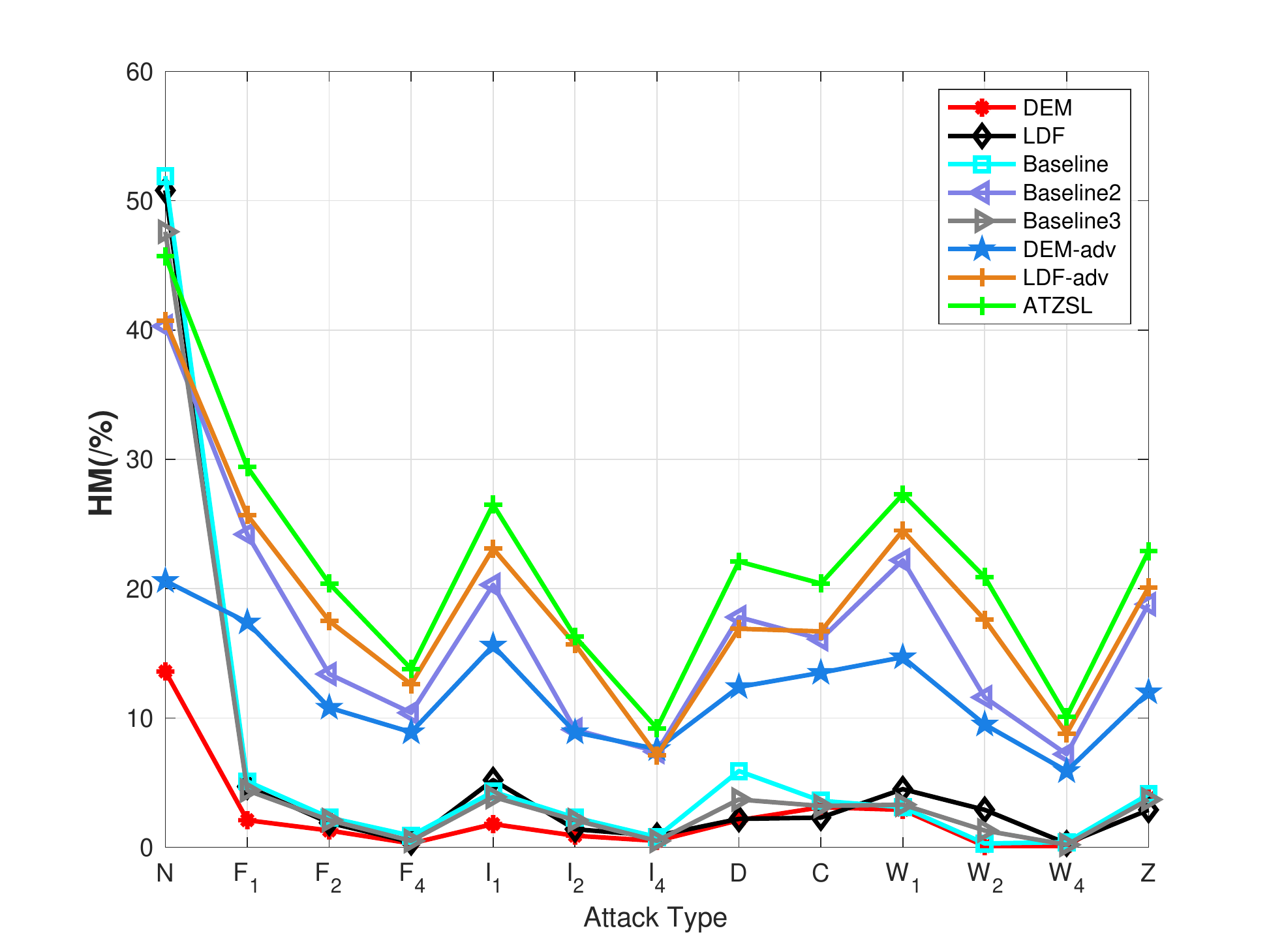}
\end{minipage}
}%
\subfigure[Attribute attack on CUB]{\label{att_gzsl_cub_aba}
\begin{minipage}[t]{0.25\linewidth}
\centering
\includegraphics[width=1.75in]{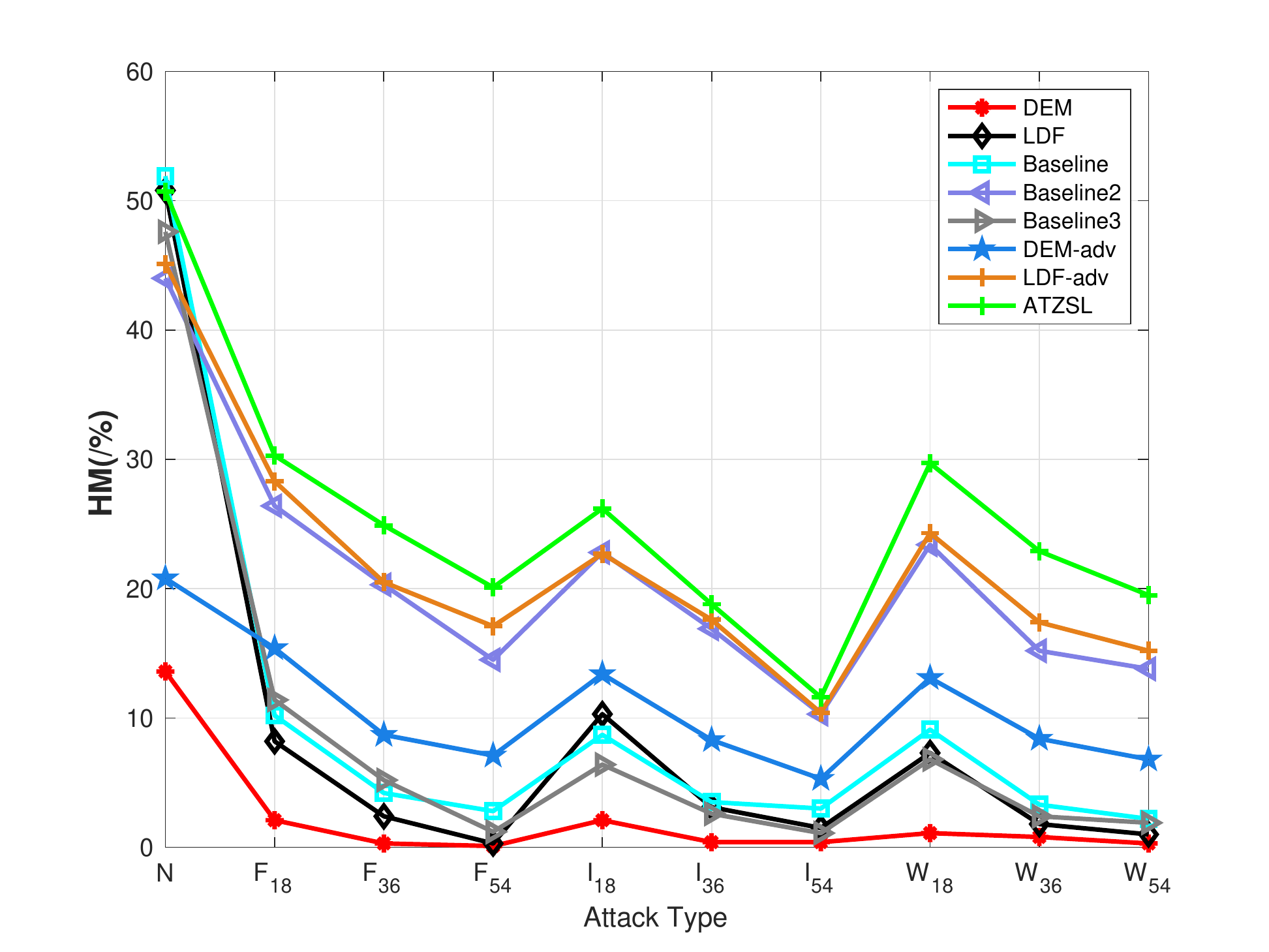}
\end{minipage}
}%
\centering
\caption{Comparison (HM:\%) of several ZSL methods with their adapted versions on two datasets under the \textbf{generalized} ZSL setting in various attack scenarios. For horizontal axis, 'N', 'F', 'I', 'D', 'C', 'W', and 'Z' represent No attack, FGSM, IFGSM, DeepFool, CW, WRM, and ZOO attack methods, respectively. The numbers with respect to 'F', 'I', and 'W' represent the attack magnitudes.}
\label{fig:GZSL_aba}
\end{figure*} 

\subsection{Ablation Study}
Actually, we have conducted a vital ablation study, namely~\textbf{Baseline} in Section~\ref{sec:szsl} and Section~\ref{sec:gzsl}. By comparison, ATZSL consistently outperforms Baseline on AWA2 and CUB datasets under both standard and generalized ZSL settings for image or attribute attacks. This just verifies that training ZSL models with adversarial samples or attributes is indeed effective. 
To further verify this fact, we additionally train the state-of-the-art ZSL methods (DEM~\cite{zhang2017learning} and LDF~\cite{li2018discriminative}) with adversarial examples, named \textbf{DEM-adv} and \textbf{LDF-adv}.
It is worth noting that GAN based ZSL methods (e.g., FGN~\cite{xian2018feature} and MC-ZSL~\cite{felix2018multi}) have not been considered since they are very complex and also cannot been adapted for image attacks.
The key to adaption is to generate adversarial images or attributes by maximizing original objective function as that in Eq.~(\ref{ATZSL}) or Eq.~(\ref{atta}).
Particularly, we need to adapt DEM~\cite{zhang2017learning} as an end-to-end model by training it with the feature extraction module (Resnet34) simultaneously.
Fig.~\ref{fig:ZSL_aba} shows the $\textrm{T1}$ performances of adapted competitors under the standard ZSL setting. 
It can be seen that for both datasets, there exists an obvious gap between DEM-adv and DEM~\cite{zhang2017learning} (resp. LDF-adv and LDF~\cite{li2018discriminative}) in terms of $\textrm{T1}$ in each attack scenario.
Likewise, under the generalized ZSL setting, we also present the performances (i.e., $\textrm{HM}$) of adapted competitors on two datasets under various attack scenarios in Fig.~\ref{fig:GZSL_aba}.
Consequently, the similarly large gap between DEM-adv and DEM~\cite{zhang2017learning} (resp. LDF-adv and LDF~\cite{li2018discriminative}) re-verifies that using adversarial data can indeed improve the robustness of ZSL models.
More importantly, as shown in Fig.~\ref{fig:ZSL} and Fig.~\ref{fig:GZSL}, under no attack scenario (denoted by `N' in horizontal axis), our \textbf{Baseline} has achieved higher performance than competitors (e.g., DEM~\cite{zhang2017learning}, LDF~\cite{li2018discriminative}, FGN~\cite{xian2018feature}, and MC-ZSL~\cite{felix2018multi}). Thus adversarially training on our baseline instead of others is more promising. 
This can be verified by an observation in Fig.~\ref{fig:ZSL_aba} and Fig.~\ref{fig:GZSL_aba} that our ATZSL consistently outperforms the two adapted ZSL methods (DEM-adv and LDF-adv).

\begin{table*}[tp!]   
\caption{Training Time (\emph{sec.}) of Each Part During An Iteration, Where the Numbers in Bold Are the Added Overhead Using Our ATZSL} \label{tab:time}
 \vspace{-0.3cm}
 \begin{center}
  \begin{tabular}{c|c|cccc|c}
    \hline 
    &  & Forward (Clean) & \textbf{Finding adversarial data}& \textbf{ Forward (Adversarial)}  & Backward   & Total  \\
    \hline
    \multirow{2}*{AWA2}  & Image attack & 22 & \textbf{226} & \textbf{22} & 103 & 378 \\
    & Attribute attack& 2 & \textbf{17}& \textbf{2}& 9& 31 \\
    \hline
    \multirow{2}*{CUB}  & Image attack &7 & \textbf{70} & \textbf{7} & 34& 119 \\
    & Attribute attack &2 &\textbf{12} & \textbf{2}& 6& 23\\
    \hline
    \end{tabular}
 \end{center}
\end{table*}

Besides, we further conduct two additional ablation studies: i) We train our ATZSL without clean samples or attributes (i.e., $\alpha=0$), named~\textbf{Baseline2}, to verify the necessity of retaining clean data. 
ii) We replace the concatenation operator in Eq.~(\ref{relation_score}) with product operator for feature maps, named~\textbf{Baseline3}, to verify the advantage of concatenation as claimed in Section~\ref{ModelFormulation}. In particular, we set $\alpha=1$ in Baseline3 for convenient implementation. 
The related results are presented in Fig.~\ref{fig:ZSL_aba} and Fig.~\ref{fig:GZSL_aba}, corresponding to the standard and the generalized ZSL settings respectively\footnote{The trade-off of each method is presented in the supplementary material.}.
It is as expected that our ATZSL performs consistently best compared with our various baselines under any attacks. Specifically, an obvious gap between the performances of ATZSL and Baseline2 demonstrates the first ablation study, while the gap between Baseline and Baseline3 manifests the second one. In essence, clean and adversarial samples play an equally important role for training a robust ZSL model.

\subsection{Complexity Analysis}
To demonstrate the efficiency of our ATZSL in Eq.~(\ref{ATZSL}) and Eq.~(\ref{atta}) for image and attribute attacks, respectively, we further provide the computational costs during model training under the standard ZSL setting (the same as the generalized ZSL setting). 
Specifically, given the network parameters $\phi$, $\varphi$, and $\phi$, the forward complexity of our model is $\mathcal{O}((2+q)d)$ to compute each pair-wise relationship score between every training example and each seen class prototype. $q$ and $d$ denote the dimensions of class prototype and image embedding, respectively.
Thus, it is actually efficient to obtain all clean scores with the complexity of $\mathcal{O}_{f}=\mathcal{O}(N_{\textrm{tr}}N_{\textrm{s}}(2+q)d)$.
Supposing $\mathcal{O}_{b}$ denotes the backward complexity of our network on all clean data, finding the worst adversarial data only relies on the gradient with respect to input but not all the variables, and thus costs must far less than $N(\mathcal{O}_{f}+\mathcal{O}_{b})$. 
In particular, $N$ is the update steps in IFGSM during training, and generally small (we set $N=9$ in our experiments).
Consequently, with an additional forward cost on the generated adversarial data, the overall cost of our algorithm is far less than $(N+2)\mathcal{O}_{f}+(N+1)\mathcal{O}_{b}$. This is practically acceptable though more than original $\mathcal{O}_{f}+\mathcal{O}_{b}$. In particular, the complexity of our baseline, i.e., $\mathcal{O}_{f}+\mathcal{O}_{b}$, is really low in ZSL compared with those methods taking far more complicated nonlinear formulation.

To present the overhead of ATZSL more intuitively, we further provide the running time of each part in our algorithm. Concretely, with 2661 MiB and 739 MiB GPU memory usage for image and attribute attacks on AWA2 dataset respectively (2717 MiB and 1207 MiB on CUB dataset), Table~\ref{tab:time} reports the physical training time during an iteration of our ATZSL. The max numbers of iteration are 300 and 2000 for image and attribute attacks, respectively. It can be concluded that the added overhead, including finding and training the worst adversarial data, can be neglected in practice to the significant accuracy improvements achieved by these adversarial data. In addition, our ATZSL, though adds some overhead, still takes comparable or even less time than GAN based ZSL methods (e.g., FGN~\cite{xian2018feature} and MC-ZSL~\cite{felix2018multi}).

\section{Conclusion}
In this work, we addressed the new problem of simultaneously achieving high robustness and transferability in zero-shot learning. For this end, we developed ATZSL: an adversarially trained zero-shot model by integrating the two goals in one unified constrained optimization framework.
Specifically, a defensive relation prediction network was designed to transfer knowledge from the seen to the unseen class domains. Meanwhile, during the training phase, we injected adversarial images or attributes into our network to transfer knowledge from the clean to the adversarial domains, thus learning a robust ZSL model. 
Extensive experiments have been carried out to endorse the effectiveness and efficiency of ATZSL by observing the following: i) compared with state-of-the-art ZSL models and their variants, our ATZSL performs consistently better on various adversarial data for zero-shot recognition task, while only loses a negligible performance on clean data; ii) under the standard or generalized ZSL setting, ATZSL can consistently remain robust against different attackers both in the visual and semantic spaces. 

We emphasize the first attempt to study robustness of ZSL creatively, which is really practically important. Moreover, ATZSL could serve as a benchmark for later scholars, though defense with adversarial data is general. 
Based on ATZSL, we are recently trying to narrow the large accuracy gap between clean and adversarial samples on unseen classes of ZSL, with multi-level distribution alignment constraint. This is a valuable issue since it is still challenging even on general classification tasks. 








\ifCLASSOPTIONcaptionsoff
  \newpage
\fi




\bibliographystyle{IEEEtran}
\bibliography{Mybib}

\end{document}